\documentclass{article}

\usepackage{iclr2026_conference,times}

\usepackage[T1]{fontenc}
\usepackage{amsthm}
\newtheorem{theorem}{Theorem}
\newtheorem{remark}{Remark}
\newtheorem{lemma}{Lemma}
\usepackage{booktabs}
\usepackage{multirow}
\usepackage{subcaption}
\usepackage{colortbl}

\usepackage{physics}
\usepackage{xcolor}
\usepackage{amsmath}
\usepackage{amssymb}
\usepackage{microtype}
\usepackage{graphicx}
\usepackage{booktabs} % for professional tables

\usepackage{hyperref}

\usepackage{algorithmic,algorithm}

\DeclareMathOperator*{\argmax}{arg\,max}

\title{Towards Understanding The Calibration Benefits of Sharpness-Aware Minimization}

\author{%
  Chengli~Tan$^{1,2}$,
  Yubo~Zhou$^{2,3}$,
  Haishan~Ye$^{2,3,\ast}$,
  Guang~Dai$^{2}$,
  Junmin~Liu$^{2,3,}$\thanks{Corresponding authors (junminliu,yehaishan@mail.xjtu.edu.cn).}
  \AND
  Zengjie~Song$^{3}$,
  Jiangshe Zhang$^{3}$, 
  Zixiang~Zhao$^{4}$, 
  Yunda~Hao$^{5}$, 
  Yong~Xu$^{1}$ \\
  \\
  {}$^1$~Northwestern Polytechnical University \quad
  {}$^2$~SGIT AI Lab, State Grid Corporation of China \\
  {}$^3$~Xi'an Jiaotong University \quad
  {}$^4$~ETH Zürich \quad
  {}$^5$~Chinese University of Hong Kong
}

\iclrfinalcopy
\begin{document}

\maketitle

\begin{abstract}
Deep neural networks have been increasingly used in safety-critical applications such as medical diagnosis and autonomous driving.
However, many studies suggest that they are prone to being poorly calibrated and have a propensity for overconfidence, which may have disastrous consequences.
In this paper, unlike standard training such as stochastic gradient descent, we show that the recently proposed sharpness-aware minimization (SAM) counteracts this tendency towards overconfidence.
The theoretical analysis suggests that SAM allows us to learn models that are already well-calibrated by implicitly maximizing the entropy of the predictive distribution.
Inspired by this finding, we further propose a variant of SAM, coined as CSAM, to ameliorate model calibration.
Extensive experiments on various datasets, including ImageNet-1K, demonstrate the benefits of SAM in reducing calibration error.
Meanwhile, CSAM performs even better than SAM and consistently achieves lower calibration error than other approaches.
\end{abstract}

\section{Introduction}
\label{sec:introduction}
While the relation between generalization and flatness is still in dispute \citep{dinh2017sharp,ramasinghe2023much,andriushchenko2023modern,wen2024sharpness}, it is empirically appreciated that under some constraints, the flatter solutions tend to generalize better \citep{hinton1993keeping,keskarlarge2017,chaudhari2019entropy,kaddour2022flat}.
From this point of view, many approaches have been proposed to bias solutions toward flat regions of the loss landscape explicitly or implicitly \citep{huangsnapshot2017,izmailov2018averaging,chaudhari2019entropy,zhang2019lookahead,wang2021eliminating,bisla2022low}, amongst which SAM \citep{foretsharpness2021} has garnered increasing attention due to its surprising effectiveness on popular tasks such as image classification \citep{chen2021vision}, language generation \citep{bahri2022sharpness}, and even physical computation \citep{xu2024perfecting}.

Different from standard training like stochastic gradient descent (SGD), SAM minimizes a perturbed loss, and each iteration is composed of two consecutive steps,
\begin{equation*}
    \tilde{\boldsymbol{\theta}}_{t} = \boldsymbol{\theta}_t + \rho \frac{\nabla L_{\Omega_t}(\boldsymbol{\theta}_t)}{\|\nabla L_{\Omega_t}(\boldsymbol{\theta}_t)\|_2},\quad \boldsymbol{\theta}_{t+1} = \boldsymbol{\theta}_t - \eta \nabla L_{\Omega_t}(\tilde{\boldsymbol{\theta}}_{t}),
\end{equation*}
where $\boldsymbol{\theta}_t\in\mathbb{R}^d$ represents the learnable parameters of the neural network at $t$-th iteration, $\eta$ is the learning rate, $\rho$ is the perturbation radius, and $L_{\Omega_t}(\cdot)$ denotes the empirical loss on a mini-batch $\Omega_t$ of the training set $S$.
This scheme constantly penalizes the gradient norm \citep{zhao2022penalizing,wen2022does,compagnoni2023sde} and significantly promotes generalization.
On the other hand, model calibration refers to how reliable the model predictions are.
Ideally, when the model is confident about its predictions, the predictions are supposed to be as accurate as possible.
This is particularly important for real-world applications such as autonomous driving \citep{chib2023recent} and medical diagnosis \citep{jiang2012calibrating}.
As an example, consider a self-driving car that uses deep neural networks to detect whether an obstruction is a pedestrian or not.
For an ill-calibrated model, when its confidence is low, it may just pass through and will not trigger emergency braking, which could cause undesired consequences.
In contrast, for a well-calibrated model, it is not certain whether the obstruction is a pedestrian or not when its confidence is low.
As a result, a more cautious decision would be made by the car to avoid an accident.
\begin{figure}[t]
\centering
\includegraphics[width=0.95\textwidth, clip, trim= 0 0 0 0]{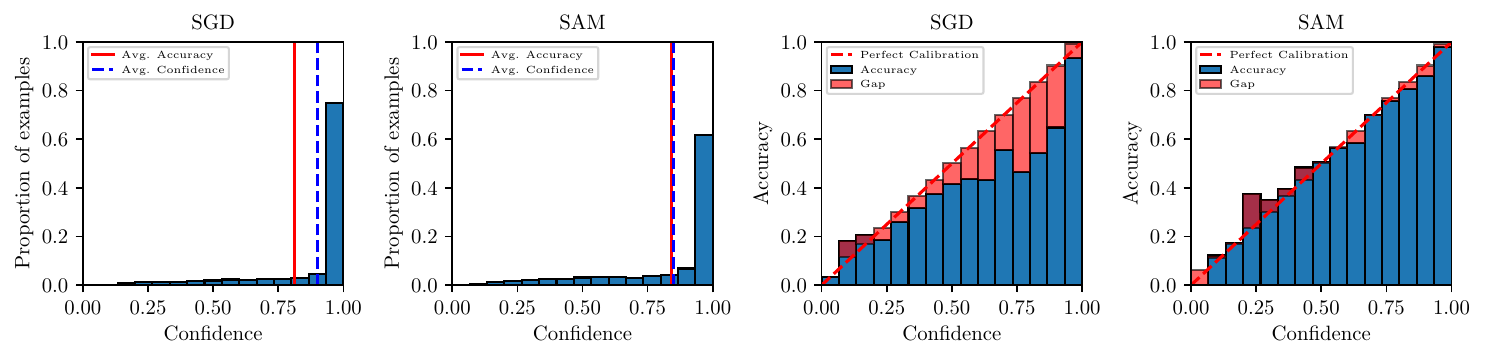}
\caption{Confidence and reliability histograms for a PyramidNet \citep{han2017deep} trained on CIFAR-100 \citep{krizhevsky2009learning} with different optimizers.
For clarity, the term \emph{confidence} here refers to the predicted probability, namely, the maximum output of the softmax layer.
}
\label{fig: calibration of sgd and sam}
\end{figure}
\par
It is known that modern neural networks such as ResNets \citep{he2016deep} and DenseNets \citep{huang2017densely} often suffer from the miscalibration problem, and this issue appears to be more serious when the network starts to overfit the training data \citep{nguyen2015deep,guo2017calibration,zhu2023rethinking}.
Since SAM is more effective in preventing overfitting \citep{foretsharpness2021}, one could anticipate that neural networks optimized by SAM may be better calibrated than by base optimizers such as SGD and AdamW \citep{loshchilov2019decoupled}.
This is illustrated in Figure \ref{fig: calibration of sgd and sam}, where a large PyramidNet \citep{han2017deep} is respectively trained on CIFAR-100 \citep{krizhevsky2009learning} with SGD and SAM.
One can easily observe that the average confidence of SAM closely matches its accuracy, while the average confidence of SGD is substantially higher than its accuracy.
This is further confirmed with a reliability diagram \citep{niculescu2005predicting}, where we plot the accuracy as a function of the confidence.
The diagram indicates that SAM is better calibrated than SGD, as the accuracy almost overlaps with the confidence along the diagonal line.
\par
While previous studies \citep{zheng2021regularizing,mollenhoffsam2023} have reported this phenomenon, the question of how SAM alleviates the miscalibration problem has not been formally investigated, and we attempt to fill this gap in this paper.
In brief, our contributions are as follows:
\begin{itemize}
    \item We provide theoretical justification for the calibration benefits of SAM, which essentially performs an implicit regularization on the negative entropy of the predictive distribution. This is similar to focal loss \citep{mukhoti2020calibrating}, but SAM calibrates models much better without compromising accuracy.
    \item We investigate how SAM performs on model calibration under distribution shift and find that SAM allows models to remain well-calibrated under different types of corruption.
    Moreover, the trick of ensembling is also useful for SAM, and compared to SGD, the improvement is more pronounced on out-of-distribution data.
    \item We develop a variant of SAM, termed CSAM, that attempts to improve model calibration further.
    By extensive experiments with a variety of network architectures and datasets, we observe that CSAM consistently performs better than SAM and surpasses other approaches that are focused on improving calibration.
    % More importantly, this superiority is retained even when models are further calibrated by post-hoc methods like temperature scaling \citep{guo2017calibration}.
\end{itemize}
The remainder of the paper is organized as follows.
We first review the related work in Section \ref{sec: related work} and then introduce some backgrounds in Section \ref{sec: preliminaries}.
After presenting the theoretical analysis of SAM and the derivation of CSAM in Section \ref{sec: theoretical analysis}, we further provide the experimental results in Section \ref{sec: experiments}.
\section{Related work}
\label{sec: related work}
In this section, we present the most relevant works on SAM and the miscalibration of deep neural networks.
\paragraph{Sharpness-aware minimization.} Because SAM is particularly effective in improving the generalization performance of realistic neural networks \citep{foretsharpness2021,chen2021vision,bahri2022sharpness}, it has received a lot of attention in recent years, and there is a surge of research along this direction.
For example, to reduce the computational overhead incurred by the additional backpropagation, some works choose to apply SAM and standard training alternatively \citep{liu2022towards,zhao2022randomized,jiang2022adaptive,tan2024sharpness}, while some other works focus on perturbing a fraction of parameters \citep{du2021efficient,mi2022make} or examples \citep{ni2022k}.
Concurrently, some researchers also attempt to further enhance the generalization performance of SAM \citep{zhang2022ga,li2023enhancing,yue2023sharpness,zhou2023imbsam}.
For example, \citet{kwon2021asam} propose ASAM to consolidate the correlation between sharpness and generalization, which might break up due to model reparameterization \citep{dinh2017sharp}.
And \citet{kim2022fisher} further propose FisherSAM to enforce that the optimization occurs on the statistical manifold induced by the Fisher information matrix.
\par
On the theoretical aspect, \citet{wen2022does,bartlett2023dynamics} prove that the largest eigenvalue of the Hessian decreases along the trajectory of SAM, a result which is quite similar to that of \citet{compagnoni2023sde} though derived from the perspective of the stochastic differential equation.
\citet{andriushchenko2022towards} propose to study the unnormalized SAM and demonstrate the implicit bias on simple diagonal neural networks.
Based on uniform stability \citep{bousquet2002stability,hardt2016train}, \citet{tan2024stabilizing} prove that SAM generalizes better than SGD on strongly convex problems, and propose a renormalization trick to mitigate the instability issue near the saddle points \citep{compagnoni2023sde,kim2023stability}.
\paragraph{Miscalibration of deep neural networks.}
In machine learning, calibration has been extensively studied \citep{platt1999probabilistic,gneiting2007probabilistic,futamiinformation2024}.
Since popular classification losses like squared error and cross-entropy (CE) are proper scoring rules \citep{gneiting2007probabilistic}, they are guaranteed to produce perfectly calibrated models at their global minimum.
However, as first disclosed by \citet{guo2017calibration}, modern neural networks suffer from serious miscalibration due to overfitting and overparameterization \citep{lakshminarayanan2017simple,thulasidasan2019mixup,wang2021rethinking,wang2023calibration}.
While \citet{minderer2021revisiting} argue that the most recent non-convolutional models like MLP-Mixer \citep{tolstikhin2021mlp} and vision transformers \citep{dosovitskiy2020image} are better calibrated, the issue of miscalibration is still prevalent in a wide spectrum of applications like data distillation \citep{zhu2023rethinking} and object detection \citep{kuzucu2025calibration}.
\par
A variety of approaches have been proposed to improve model calibration.
In the training-time calibration, for example, an intuitive idea is to penalize overconfidence, either explicitly via entropy-based regularization \citep{pereyra2017regularizing} and label smoothing \citep{muller2019does} or implicitly using focal loss (FL) \citep{mukhoti2020calibrating,tao2023dual}.
However, as pointed out by previous works \cite{wang2021rethinking,singh2021deep}, the penalty of confident outputs may suppress the potential improvement in the post-hoc calibration phase.
On the other hand, post-hoc calibration addresses the miscalibration problem by appending a post-processing step to the training phase and typically requires a hold-out validation set for hyperparameter tuning.
Popular post-hoc methods include non-parametric calibration methods---histogram binning \citep{zadrozny2001obtaining}  and isotonic regression \citep{zadrozny2002transforming}, and parametric methods like Bayesian binning \citep{naeini2015obtaining} and Platt scaling \citep{platt1999probabilistic}.
Out of them, Platt scaling-based approaches such as temperature scaling \citep{guo2017calibration} and Dirichlet calibration \citep{kull2019beyond} are more frequently used due to their low complexity and efficiency.

\section{Preliminaries}
\label{sec: preliminaries}
In this section, we first introduce one measure of model calibration that we use throughout, and then briefly recap the difference between SAM and SGD.
Without loss of generality, we consider the multi-class classification problem where a categorical variable $Y\in\{1, \dots, K\}$ is predicted when an input variable $X$ is observed.
And we further assume that the training set $S$ contains $n$ examples $\{z_i=(x_i, y_i)\}_{i=1}^n$ that are \emph{i.i.d.}~sampled from an unknown data distribution $\mathcal{D}$.
For a deep neural network parameterized by $\boldsymbol{\theta}\in \mathbb{R}^d$, we naturally obtain a predictor $f_{\boldsymbol{\theta}}$ that maps the features $X$ to a categorical distribution over $K$ labels, which we denote it by $f_{\boldsymbol{\theta}}(X)$ that belongs to a $(K-1)$-dimensional simplex $\Delta=\{\mathbf{p}\in[0, 1]^K|\sum_{y=1}^K\mathbf{p}_y=1\}$.
Then, $\hat{y}\triangleq\argmax_{1\leq y\leq K}\mathbf{p}_y$ is the predicted label.
% and $\hat{p}\triangleq\max_{1\leq y\leq K}\mathbf{p}_y$ is the associated confidence.
 \subsection{Expected calibration error}
 A model is well-calibrated if the confidence truthfully recovers the probability of correctness.
 That is, if we gather all data points for which the model predicts $\mathbf{p}_y=0.8$, we expect that 80\% of them should take on the label $y$.
 Mathematically, we refer to a model as well-calibrated \citep{brocker2009reliability} if 
 \begin{equation*}
     P(Y=y| f_{\boldsymbol{\theta}}(X) = \mathbf{p}) = \mathbf{p}_y, \quad \forall~ \mathbf{p}\in \Delta.
 \end{equation*}
 In practice, however, we will focus on the top-label calibration \citep{guo2017calibration} that requires the above equation to hold only for the most likely label, namely,
 \begin{equation*}
     P(Y=\hat{y} |\max_{1\leq y\leq K}\mathbf{p}_y = \hat{p}) = \hat{p}, \quad \forall~ \hat{p}\in[0, 1].
 \end{equation*}
 Expected calibration error (ECE) is the most commonly used metric to measure the degree of miscalibration, which quantifies the expected difference between two sides of the above equation as follows
 \begin{equation*}
     \mathbb{E}\left[\left|\hat{p} - P(Y=\hat{y} |\max_{1\leq y\leq K}\mathbf{p}_y = \hat{p})\right|\right].
 \end{equation*}
% For perfectly calibrated models, ECE is equal to zero.
 In practice, due to finite examples, it works by firstly grouping all examples, say, $\{z_i = (x_i, y_i)\}_{i=1}^{n}$, into $M$ bins $B_1, \ldots, B_M$ based on their top confidence scores.
 Next, we compute in each bin $B_i$ the average confidence $\operatorname{conf}(B_i) = 1/|B_i|\sum_{z_j\in B_i} \max f_{\boldsymbol{\theta}}(x_j)$ and the average accuracy $\operatorname{acc}(B_i) = 1/|B_i|\sum_{z_j\in B_i} \mathbb{I}[y_j=\argmax  f_{\boldsymbol{\theta}}(x_j)]$, where $\mathbb{I}[\cdot]$ is the indicator function.
 Then, we can obtain an estimator by averaging over the bins
 \begin{equation*}
     \widehat{\operatorname{ECE}} = \sum_{i=1}^{M} \frac{|B_i|}{n} \left|\operatorname{acc}(B_i) - \operatorname{conf}(B_i)\right|.
 \end{equation*}
 \subsection{Sharpness-aware minimization}
The intuitive idea of SAM \citep{foretsharpness2021} is to improve generalization by constantly minimizing the solution sharpness during training.
To this end, instead of minimizing the loss at the current point, it minimizes the worst-case loss within its neighborhood.
Mathematically, it is equivalent to solving the following optimization problem,
\begin{equation*}
    \min_{\boldsymbol{\theta}\in\mathbb{R}^d} \max_{\|\boldsymbol{\varepsilon}\|_2 \leq \rho} L_S(\boldsymbol{\theta}+\boldsymbol{\varepsilon}),
\end{equation*}
where $\boldsymbol{\varepsilon}\in\mathbb{R}^d$ is a perturbation vector whose norm is bounded by the perturbation radius $\rho>0$.
It is not easy to solve this minimax problem explicitly.
But, after a simple Taylor approximation, we observe that
\begin{align*}
    \boldsymbol{\varepsilon}^\ast & \triangleq \argmax_{\|\boldsymbol{\varepsilon}\|_2 \leq \rho} L_S(\boldsymbol{\theta}+\boldsymbol{\varepsilon}) \\
    &\approx \argmax_{\|\boldsymbol{\varepsilon}\|_2 \leq \rho}L_S(\boldsymbol{\theta}) + \boldsymbol{\varepsilon}^T \nabla L_S(\boldsymbol{\theta}) 
    = \rho \frac{\nabla L_S(\boldsymbol{\theta})}{\|\nabla L_S(\boldsymbol{\theta})\|_2}.
\end{align*}
This suggests that, as opposed to SGD, we first need to do an extra gradient backpropagation to estimate the perturbed vector $\boldsymbol{\varepsilon}^\ast$.
Therefore, SAM actually consists of two consecutive steps at each iteration,
\begin{equation*}
    \tilde{\boldsymbol{\theta}}_{t} = \boldsymbol{\theta}_t + \rho \frac{\nabla L_{\Omega_t}(\boldsymbol{\theta}_t)}{\|\nabla L_{\Omega_t}(\boldsymbol{\theta}_t)\|_2},\quad \boldsymbol{\theta}_{t+1} = \boldsymbol{\theta}_t - \eta \nabla L_{\Omega_t}(\tilde{\boldsymbol{\theta}}_{t}),
\end{equation*}
where $\Omega_t$ denotes a random mini-batch of $S$.
We note that the same $\Omega_t$ is used for the ascent and descent steps, and a smaller $\Omega_t$ is preferred in practice for better generalization \citep{foretsharpness2021,andriushchenko2022towards}.
% \begin{figure}[t]
% 	\centering
% 	\includegraphics[width=0.4\linewidth]{figs/monotone_function.pdf}
% 	\caption{Value of the coefficient $\lambda$ as a monotone function of the perturbation radius $\rho$ and the predicted probability associated with the true label $\tilde{\mathbf{p}}_y$.}
% 	\label{fig: monotone function}
% \end{figure}
\begin{figure}[t]
\centering
\begin{subfigure}[b]{0.31\linewidth}
    \centering
    \includegraphics[width=\textwidth, clip, trim= 0 0 0 0]{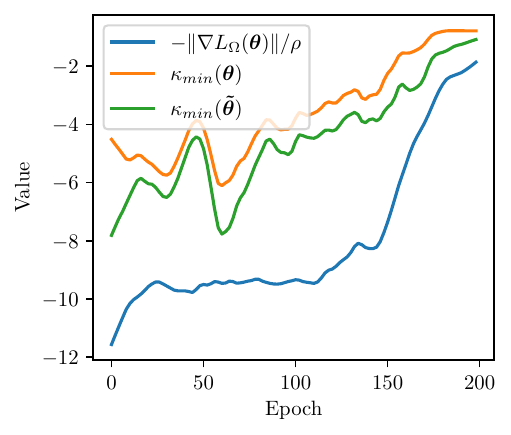}
    \caption{}
\end{subfigure}
\begin{subfigure}[b]{0.31\linewidth}
    \centering
    \includegraphics[width=\textwidth, clip, trim= 0 0 0 0]{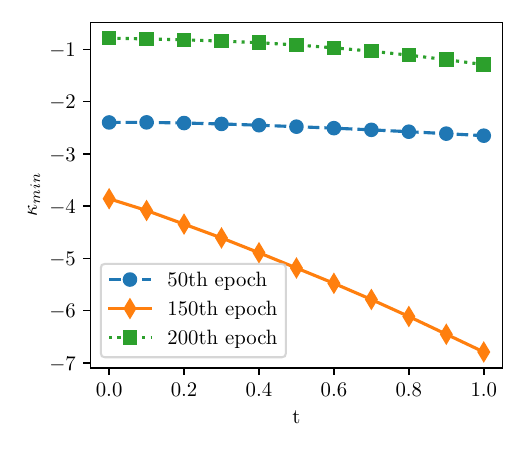}
    \caption{}
\end{subfigure}
\begin{subfigure}[b]{0.31\linewidth}
    \centering
    \includegraphics[width=\textwidth, clip, trim= 0 0 0 0]{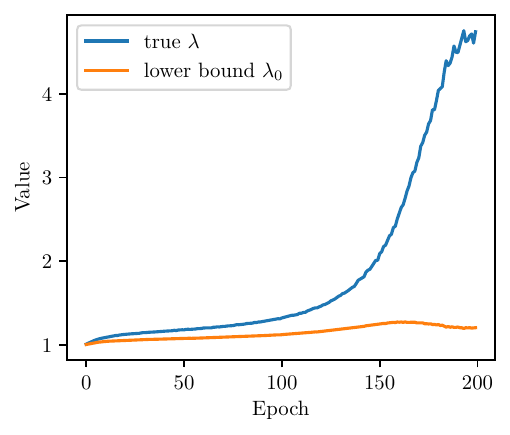}
    \caption{}
\end{subfigure}
\caption{To verify whether the boundedness assumption of $\kappa_{min}$ (see Lemmas \ref{lemma: one-sam} and \ref{lemma: m-sam}) holds for realistic neural networks, we trained a ResNet-56 on CIFAR-10 using a constant $\rho=0.05$. (a) compares the value of $\kappa_{min}$ at the two endpoints $\boldsymbol{\theta}$ and $\tilde{\boldsymbol{\theta}}$ throughout the training process. (b) further illustrates how $\kappa_{min}$ evolves along the path from $\boldsymbol{\theta} (t=0)$ to $\tilde{\boldsymbol{\theta}} (t=1)$ at 50th, 150th, and 200th epoch, respectively. (c) records how the true coefficient $\lambda$ and its lower bound $\lambda_0$ (see Equation \ref{eq: ratio lower bound}) vary during training.
Notice that similar results for the realistic ImageNet-1K dataset can be found in Figure \ref{fig: vit monotone function}.
}
\label{fig: monotone function}
\end{figure}
\section{Methodology}
\label{sec: theoretical analysis}
In this section, we first show that SAM is bound to prevent deep neural networks from producing overconfident predictions.
As in previous studies \citep{guo2017calibration,minderer2021revisiting,wang2021rethinking}, we focus on the most widely used cross-entropy (CE) loss in the classification problem, which for an example $z=(x, y)$ is defined as $\ell_{\boldsymbol{\theta}}(z)=-\log \mathbf{p}_{y}$ in one-hot encoding.
The analysis is straightforward, and all proofs are deferred to Appendix \ref{appendix: missing proofs} for clarity.
Towards the end of this section, we also develop a variant of SAM to improve its calibration performance.
\subsection{Theoretical analysis}
Let $\mathbf{p}_{y}=[f_{\boldsymbol{\theta}}(x)]_y$ and $\tilde{\mathbf{p}}_{y}=[f_{\tilde{\boldsymbol{\theta}}}(x)]_y$ denote the confidence on the true label $y$ conditioned on the current weight $\boldsymbol{\theta}$ and the perturbed weight $\tilde{\boldsymbol{\theta}}$, respectively.
When the mini-batch is 1, namely, every step we sample one example only to estimate the true gradient, the following lemma suggests that $\tilde{\mathbf{p}}_{y}$ can be consistently smaller than $\mathbf{p}_{y}$ during training.
\begin{lemma}[1-SAM version]
\label{lemma: one-sam}
    Assume that at each step, the gradient $\nabla_{\boldsymbol{\theta}}\ell(z) \neq \mathbf{0}$ and there always exists some $\rho>0$ such that the smallest eigenvalue of the Hessian $\kappa_{min}(\nabla^2 \ell_{\boldsymbol{\theta}^\prime}(z))\geq-\|\nabla_{\boldsymbol{\theta}}\ell(z)\|/\rho$ holds for all  $\boldsymbol{\theta}^\prime = (1-t)\boldsymbol{\theta}+t\tilde{\boldsymbol{\theta}}$, $t\in[0, 1]$.
    Then, given $\mathbf{p}_{y}$, $\tilde{\mathbf{p}}_{y}$ defined as above, we have $\tilde{\mathbf{p}}_{y} \leq e^{-\rho\|\nabla_{\boldsymbol{\theta}}\ell(z)\|/2}\mathbf{p}_{y}$.
\end{lemma}
Actually, the boundedness of $\kappa_{min}$ at $\boldsymbol{\theta}$ can be easily verified along the optimization trajectory \citep[Section 6.2]{zhou2021towards}.
However, it should be noted that the inequality does not necessarily hold for all $\boldsymbol{\theta}^\prime$.
But if we vary $\rho$ accordingly at each step ($\rho\to0$ in the worst case), the validity of the inequality can be assured because we are always ascending along the gradient direction.  
This lemma shows that $\tilde{\mathbf{p}}_y$, the probability of the perturbed network assigned to the true label, exponentially decays with the perturbation radius $\rho$ and the gradient norm $\|\nabla_{\boldsymbol{\theta}}\ell(z)\|$.
\begin{remark}
    A similar result for the mini-batch SAM is also developed in Lemma \ref{lemma: m-sam}.
    Notice that varying $\rho$ at every step is quite different from the practical setting, in which we often use a constant $\rho$ instead.
    Luckily, as Figure \ref{fig: monotone function}(a) suggests, the boundedness assumption of $\kappa_{min}$ can be validated for the constant $\rho$ over mini-batch SAM.
    Surprisingly, we also find that $\kappa_{min}$ linearly decreases along $\boldsymbol{\theta}$ to $\tilde{\boldsymbol{\theta}}$ (see Figure \ref{fig: monotone function}(b)), suggesting that the boundedness assumption can be simplified to requiring $\kappa_{min}(\nabla^2 \ell_{\tilde{\boldsymbol{\theta}}}(z))\geq-\|\nabla_{\boldsymbol{\theta}}\ell(z)\|/\rho$ only.
    This finding further reveals why a large value of $\rho$ is not preferred because the boundedness assumption can be easily violated in that case.
\end{remark}
\begin{remark}
    It should be highlighted that the gradient norm $\|\nabla_{\boldsymbol{\theta}}\ell(z)\|$ also plays a critical role in determining $\tilde{\mathbf{p}}_{y}$.
    Lemma \ref{lemma: one-sam} indicates that the SAM optimizer is more effective for larger gradient norm, while simultaneously allowing us to choose a relatively large $\rho$. 
    This finding is aligned with the observation that SAM is particularly effective in training ViT models with AdamW, which eventually improves more than 5\% accuracy on ImageNet-1K using a large $\rho$ \citep{chen2021vision}.
\end{remark}
Under Lemma \ref{lemma: one-sam}, we show that minimizing the perturbed loss $\ell_{\tilde{\boldsymbol{\theta}}}(z)$ has the same effect as adding a maximum-entropy regularizer to $\ell_{\boldsymbol{\theta}}(z)$ as focal loss (FL) \citep[Section 4]{mukhoti2020calibrating}.
\begin{theorem}[1-SAM version]
\label{theorem: one-sam}
Let $\lambda = (1-\tilde{\mathbf{p}}_y)/(1-\mathbf{p}_y)$, the following inequality holds
\begin{equation}
\label{eq:theorem one sam}
    \ell_{\tilde{\boldsymbol{\theta}}}(z) \geq \ell_{\boldsymbol{\theta}}(z) - \lambda H(\mathbf{p}_y) + H(\tilde{\mathbf{p}}_y),
\end{equation}
where $H(p) = -p\log p - (1-p)\log(1-p)$ is the binary entropy function.
\end{theorem}
According to Lemma \ref{lemma: one-sam}, we know that the coefficient $\lambda$ is larger than 1, which implies that minimizing $\ell_{\tilde{\boldsymbol{\theta}}}(z)$ implicitly puts more emphasis on maximizing $H(\mathbf{p}_y)$ in contrast to minimizing $H(\tilde{\mathbf{p}}_y)$.
That is, SAM forces $\mathbf{p}_y$ to be smaller when it approaches 1 and to be larger when it is near 0.
Moreover, when replacing $\mathbf{p}_{y}$ with $e^{\rho\|\nabla_{\boldsymbol{\theta}}\ell(z)\|/2}\tilde{\mathbf{p}}_{y}$, we have
\begin{equation}
\label{eq: ratio lower bound}
	\lambda \geq \lambda_0 = \frac{1-\tilde{\mathbf{p}}_{y}}{1-e^{\rho\|\nabla_{\boldsymbol{\theta}}\ell(z)\|/2}\tilde{\mathbf{p}}_{y}}.
\end{equation}
We note that the penalty on maximizing $H(\mathbf{p}_y)$ is stronger at the terminal phase of training than at the initial phase (see Figure \ref{fig: monotone function}(c)).
Since model architecture is also a major determinant of model calibration \citep{minderer2021revisiting}, it suggests that SAM could calibrate better for model architectures that are seriously overconfident.
%, suggesting that SAM is able to modulate the logits like temperature scaling \citep{guo2017calibration}.
%This is also reminiscent of a recent observation that SAM adaptively suppresses the well-learned features while giving more opportunity to the remaining features to be learned \citep{springer2024sharpness}.
%\subsection{Extension to $m$-SAM}
\par
In practice, as suggested by \citep{foretsharpness2021,andriushchenko2022towards}, we attempt to minimize the so-called $m$-sharpness to achieve the largest performance increment.
Different from $1$-SAM, in every step we determine the ascent direction using the gradient averaged over a mini-batch $\Omega$ of $m$ examples.
As a result, the gradient corresponding to one example $\nabla \ell_{\boldsymbol{\theta}}(z)$ is not promised to align well with the mini-batch gradient $\nabla L_{\Omega}(\boldsymbol{\theta})\triangleq 1/m \sum_{i=1}^m \nabla \ell_{\boldsymbol{\theta}}(z_i)$.
Therefore, the relation $\tilde{\mathbf{p}}_{y_i} \leq \mathbf{p}_{y_i}$ does not necessarily hold for all $z_i\in\Omega$.
However, when both of them are taken into account, we do have a result similar to Lemma \ref{lemma: one-sam} as follows.
\begin{lemma}[m-SAM version]
\label{lemma: m-sam}
Assume that at each step, the gradient $\nabla L_{\Omega}(\boldsymbol{\theta}) \neq \mathbf{0}$ and there always exists some $\rho>0$ such that the smallest eigenvalue of the Hessian $\kappa_{min}(\nabla^2 L_{\Omega}(\boldsymbol{\theta}^\prime))\geq-\|\nabla L_{\Omega}(\boldsymbol{\theta})\|/\rho$ holds for all  $\boldsymbol{\theta}^\prime = (1-t)\boldsymbol{\theta}+t\tilde{\boldsymbol{\theta}}$, $t\in[0, 1]$.
Denote $\mathbf{p}_{y} = \left(\prod_{i=1}^m\mathbf{p}_{y_i}\right)^{1/m}$ and $\tilde{\mathbf{p}}_{y} = \left(\prod_{i=1}^m\tilde{\mathbf{p}}_{y_i}\right)^{1/m}$, respectively.
Then, we have $\tilde{\mathbf{p}}_{y} \leq e^{-\rho\|L_{\Omega}(\boldsymbol{\theta})\|/2}\mathbf{p}_{y}$.
\end{lemma}
The proof is straightforward, and accordingly, we have the following result.
\begin{theorem}[m-SAM version]
\label{theorem: m-sam}
    Let $\mathbf{p}_{y}$ and $\tilde{\mathbf{p}}_{y}$ defined as above. Then, it follows that
    \begin{equation}
        L_{\Omega}(\tilde{\boldsymbol{\theta}}) \geq L_{\Omega}(\boldsymbol{\theta}) - \lambda H(\mathbf{p}_y) + H(\tilde{\mathbf{p}}_y),
    \end{equation}
    where $\lambda = (1-\tilde{\mathbf{p}}_y)/(1-\mathbf{p}_y)$.
\end{theorem}
This theorem is similar to Theorem \ref{theorem: one-sam}, albeit $\mathbf{p}_y$ is the geometric mean of the predicted probabilities.
But it is enough to make sure that $m$-SAM prevents models from producing overconfident predictions as well.
\subsection{Improving SAM towards better calibration}
As shown in Figure \ref{fig: monotone function}(c), we notice that SAM primarily starts to penalize the predictive distribution at the late stages of training where $\tilde{\mathbf{p}}_y$ is high. 
Therefore, we propose to suppress the contribution of the over-confident examples so that their predictive probability $\tilde{\mathbf{p}}_y$ is virtually higher.
That is, we can redefine the per-example loss function for the outer loop of SAM as follows:
\begin{equation}
	\label{eq: refined loss function}
	\tilde{\ell}_{\tilde{\boldsymbol{\theta}}}(z) = \begin{cases}
		-\log \tilde{\mathbf{p}}_{y}, &  \mathrm{if}\ \tilde{\mathbf{p}}_{y}\leq 1/2, \\
		- \left(1 + \tilde{\mathbf{p}}_{y}\right)^{-\gamma}\log \tilde{\mathbf{p}}_{y}, & \mathrm{otherwise},
	\end{cases}
\end{equation}
where $0\leq\gamma\leq 2$ is a hyperparameter.
It is trivial to recover the standard SAM when $\gamma=0$.
Actually, the following result suggests that the modified loss function $\tilde{\ell}_{\tilde{\boldsymbol{\theta}}}(z)$ enforces SAM to penalize the predictive distribution of over-confident examples.
\begin{theorem}
	\label{theorem: csam}
	Let Lemma \ref{lemma: one-sam} hold and $\lambda = (1-\tilde{\mathbf{p}}_y)/(1-\mathbf{p}_y)$, for all $\tilde{\mathbf{p}}_y>1/2$, the following inequality holds
	\begin{equation}
		\tilde{\ell}_{\tilde{\boldsymbol{\theta}}}(z) \geq \ell_{\boldsymbol{\theta}}(z) - \lambda H(\mathbf{p}_y) + (1-\gamma/2)H(\tilde{\mathbf{p}}_y),
	\end{equation}
	where $H(p) = -p\log p - (1-p)\log(1-p)$ is the binary entropy function.
\end{theorem}
Slightly different from Theorem \ref{theorem: one-sam}, here it brings a coefficient before $H(\tilde{\mathbf{p}}_y)$, which suggests that the implicit penalty on $H(\mathbf{p}_y)$ is stronger if $(1-\gamma/2)>0$.
Meanwhile,  we also require that $\gamma\leq 2$ so that the optimization process is always biased towards decreasing $\ell_{\tilde{\boldsymbol{\theta}}}(z)$ as in SAM.
Note that this argument is also valid for $m$-SAM as it increases the geometric mean as well.
For notational convenience, we will refer to this variant of SAM as Calibrated SAM (CSAM) in the sequel, and its pseudocode is summarized in Algorithm \ref{algo: csam} (see Appendix \ref{appendix: missing proofs}).
\begin{figure}[t]
\centering
\begin{subfigure}[b]{0.24\linewidth}
    \centering
    \includegraphics[width=\textwidth, clip, trim= 0 0 0 0]{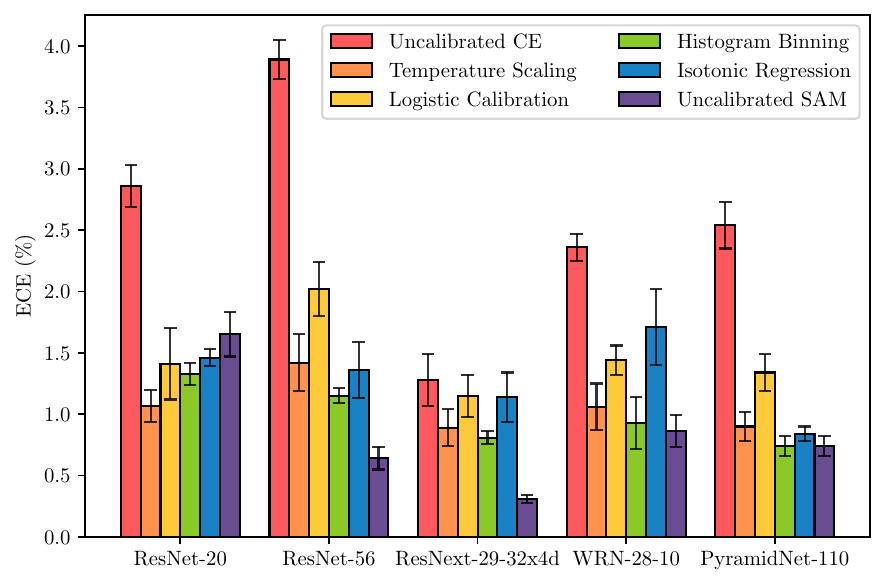}
    \caption{CIFAR-10}
\end{subfigure}
\begin{subfigure}[b]{0.24\linewidth}
    \centering
    \includegraphics[width=\textwidth, clip, trim= 0 0 0 0]{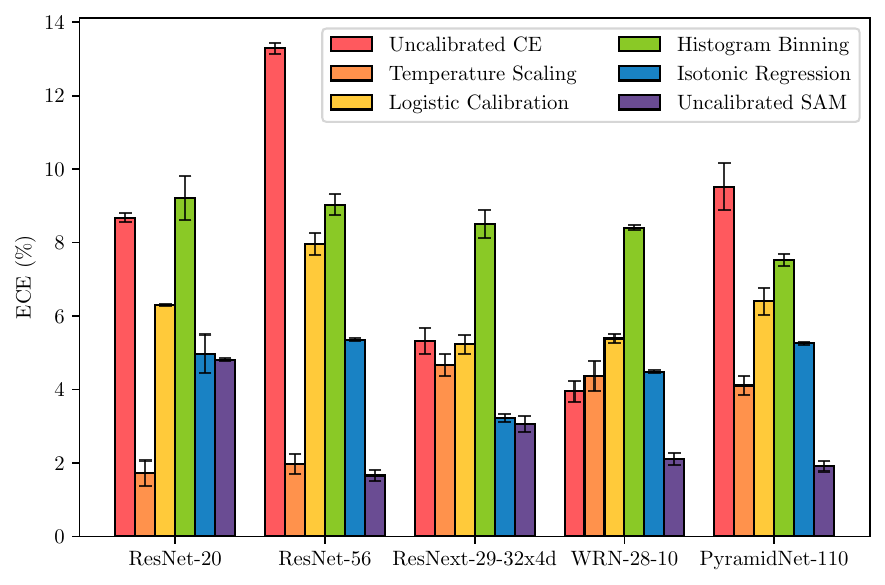}
    \caption{CIFAR-100}
\end{subfigure}
\begin{subfigure}[b]{0.24\linewidth}
    \centering
    \includegraphics[width=\textwidth, clip, trim= 0 0 0 0]{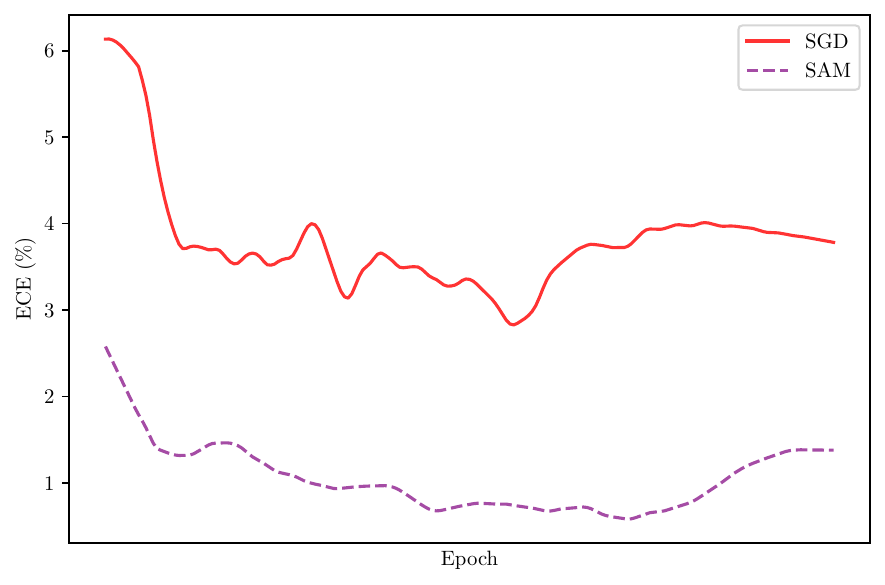}
    \caption{ECE evolution}
\end{subfigure}
\begin{subfigure}[b]{0.24\linewidth}
    \centering
    \includegraphics[width=\textwidth, clip, trim= 0 0 0 0]{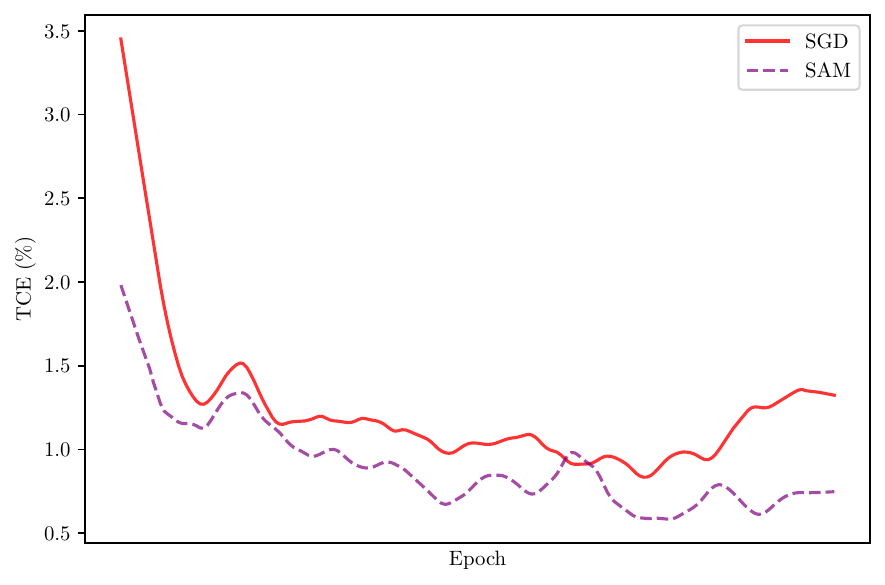}
    \caption{TCE evolution}
\end{subfigure}
\caption{(a)-(b) display the calibration performance of SAM and SGD (after various post-hoc processing) on CIFAR-10/100 datasets. (c)-(d) report the variation of ECE and TCE (namely, ECE after temperature scaling) during training.}
\label{fig: ece after post-hoc methods}
\end{figure}
\section{Experiments}
\label{sec: experiments}
In this section, we present the experimental results.
We begin with the standard benchmarks showing that SAM significantly calibrates better than SGD.
We further demonstrate on datasets including ImageNet-1K \citep{deng2009imagenet} that this calibration benefit is not limited to the in-distribution (ID) data, but also translates to the out-of-distribution (OOD) data.
At last, we compare the proposed CSAM and SAM against a variety of baselines that attempt to reduce miscalibration.
The results suggest that SAM is competitive and even superior to these approaches in many cases. 
More surprisingly, our proposed CSAM consistently outperforms SAM and achieves the lowest calibration error out of all baselines without deteriorating the generalization performance.
%We consider first the popular classification tasks on in-distribution data and then investigate the calibration performance of SAM under distribution shift.
\subsection{SAM attains a lower calibration error than SGD}
\label{sec: sam decreases ece on classification tasks}
As a starting point, we first evaluate how SAM differs from SGD on the classical benchmarks for classification.
The loss function defaults to be the standard cross-entropy (CE) loss, and we train several neural networks, including ResNets \citep{he2016deep}, Wide ResNets \citep{zagoruyko2016wide}, and PyramidNets \citep{han2017deep} to classify CIFAR-10/100 \citep{krizhevsky2009learning}.
As in common practice, we split the data into the train, validation, and test subsets so that the same validation subset is used for hyperparameter tuning and post-hoc calibration.
Without further specification, the optimizer is SGD with momentum 0.9, and the learning rate is scheduled in a cosine decay \citep{loshchilov2016sgdr}.
To conduct a fair comparison, we first make a grid search of learning rate and weight decay coefficient on the model trained with SGD, and then apply them to SAM.
The perturbation radius $\rho$ is 0.05 for CIFAR-10 and 0.2 for CIFAR-100 (see Appendix \ref{sec:Effects of Perturbation Radius} for more discussion on the effect of $\rho$ on calibration).
\par
As illustrated in Figure \ref{fig: ece after post-hoc methods}(a)-(b), the ECE of SGD (\textcolor{red}{red} bar) is always much higher than the ECE of SAM (\textcolor{purple}{purple} bar).
This is more pronounced for ResNet-56 on CIFAR-10/100, where the ECE of SGD is approximately six times larger than the ECE of SAM.
More surprisingly, we further observe that the uncalibrated ECE of SAM is generally smaller than the calibrated ECE of SGD by calibration methods such as temperature scaling \citep{guo2017calibration} and isotonic regression \citep{zadrozny2002transforming}.
This indicates that SAM by itself tends to generate accurate and reliable predictions.
Furthermore, as shown in Figure \ref{fig: ece after post-hoc methods}(c)-(d), the superiority of SAM is persistent across the full training process.
And the reduction of ECE is more pronounced than TCE since SAM has already suppressed the over-confident outputs during training, and temperature scaling is thus not as effective as in SGD.
\subsection{Model calibration under distribution shift}
\label{sec: model calibration under distribution shift}
It is important for safety-critical applications that the model not only produces reliable predictions for the in-distribution data but also is robust enough when there exists a distribution shift between the training data and the test data.
For this purpose, we first train ResNet-18 on CIFAR-10 using vanilla SGD and SAM, and then evaluate its performance on other datasets, including SVHN \citep{netzer2011reading}, CIFAR-10/100-C \citep{hendrycks2019robustness}.
To enhance model uncertainty, we further encapsulate them with MC-Dropout \citep{gal2016dropout} and Ensemble \citep{ovadia2019can}.
Table \ref{tab: cifar10 distribution shift} shows that model ensembling and MC-Dropout both can reduce ECE for SGD, SAM, and CSAM, but their gap is still significant---ECE of SGD approximately remains two times larger than ECE of SAM.
This is different from their behavior on test accuracy, for example, SGD almost generalizes as well as SAM with Ensemble.
On the other hand, it should be highlighted that SAM generalizes much better than SGD on OOD data.
And Ensemble also works well under this scenario.
An unexpected finding is that MC-Dropout hurts both optimizers' performance on OOD data and is more evident for SAM.
One possible explanation is that the fusion of Dropout and SAM adversely increases model uncertainty, which, as a result, impedes generalization.
\par
Next, we train models on the clean ImageNet-1K dataset and then assess the calibration performance of SAM on the ImageNet-C \citep{hendrycks2019robustness} dataset, which consists of images that have been modified with several synthetic corruptions at five different severities.
Following \citet{minderer2021revisiting}, we reserve 20\% of the ImageNet-1K validation set for early stopping and temperature scaling.
Moreover, we also exclude the corresponding corrupted images in ImageNet-C that are created from ImageNet-1K at the evaluation phase.
We train one ResNet and two vision transformers (ViTs) \citep{dosovitskiy2020image} on ImageNet-1K for 100 epochs and 300 epochs.
The base optimizers are SGD and AdamW, and a cosine learning rate scheduler is used in all runs.
As in previous studies \citep{foretsharpness2021,chen2021vision}, the perturbation radius $\rho$ for ResNet and ViT is 0.05 and 0.2.
\begin{table}[t]
\centering
\caption{Model performance on OOD data. The base model is ResNet-18 trained on CIFAR-10. The size of MC-Dropout and Ensemble is 5.
}
\label{tab: cifar10 distribution shift}
\resizebox{\textwidth}{!}{
\begin{tabular}{llccccc}
\toprule
& & \multicolumn{2}{c}{\textbf{ID Metrics}}& \multicolumn{3}{c}{\textbf{OOD AUROC $\uparrow$}}\\ \cmidrule{3-7}   
& & Test Acc $\uparrow$ & ECE $\downarrow$    & SVHN         & CIFAR10-C    & CIFAR100-C   \\ \midrule
  & Vanilla& $89.18 \pm 0.26$           & $5.76 \pm 0.43$          & $83.94 \pm 0.96$          & $62.26 \pm 4.46$          & $83.26 \pm 0.71$          \\ 
 SGD & MC-Dropout  & $89.13 \pm 0.18$           & $4.39 \pm 0.27$          & $84.11 \pm 0.69$          & $57.09 \pm 2.81$          & $82.11 \pm 0.82$          \\  
 &Ensemble    & $90.88 \pm 0.11$           & $1.84 \pm 0.22$          & $86.41 \pm 0.36$          & $63.39 \pm 4.72$          & $85.81 \pm 0.17$          \\ \midrule 
  &Vanilla& $90.01 \pm 0.23$           & $3.24 \pm 0.39$          & $86.38 \pm 0.39$          & $63.32 \pm 4.77$          & $84.83 \pm 0.83$          \\  
 SAM & MC-Dropout  & $89.49 \pm 0.33$           & $2.21 \pm 0.41$          & $83.02 \pm 0.35$          & $56.11 \pm 2.50$          & $80.96 \pm 0.55$          \\
 &  Ensemble    & $91.16 \pm 0.14$  & $1.09 \pm 0.22$ & $88.05 \pm 0.21$ & $64.03 \pm 4.95$ & $86.84 \pm 0.75$ \\ \midrule
 &Vanilla& $89.95 \pm 0.16$           & $2.55 \pm 0.24$          & $85.98 \pm 0.42$          & $63.49 \pm 4.74$          & $84.87 \pm 0.86$          \\  
  CSAM & MC-Dropout  & $89.57 \pm 0.21$           & $1.52 \pm 0.21$          & $82.82 \pm 0.31$          & $56.05 \pm 2.48$          & $80.70 \pm 0.55$          \\
 &  Ensemble    & $\mathbf{91.22 \pm 0.17}$  & $\mathbf{0.86 \pm 0.17}$ & $\mathbf{88.21 \pm 0.12}$ & $\mathbf{64.17 \pm 4.91}$ & $\mathbf{86.92 \pm 0.70}$ \\
 \bottomrule
\end{tabular}
}
\end{table}

\begin{table}[t]
\centering
\caption{
Results on the ImageNet-1K dataset.
Slightly different from the custom setting, we reserve 20\% of the ImageNet-1K validation set as a new validation set for early stopping and temperature scaling, and the remaining images therefore constitute a test set.
Both metrics (TCE is short for ECE calibrated by temperature scaling, and AdaECE is adaptive ECE) are evaluated on the test set.
}
\label{tab:imagenet result}
\resizebox{\textwidth}{!}{
\begin{tabular}{ccccccccccccc}
\toprule
 & & \multicolumn{5}{c}{\textbf{ID Metrics}}& \multicolumn{6}{c}{\textbf{OOD Metrics}}\\
\cmidrule{3-13}
 & &Test Acc $\uparrow$ & ECE $\downarrow$ & TCE $\downarrow$ & AdaECE $\downarrow$ & AUROC $\uparrow$ & \multicolumn{3}{c}{Test Acc (1/2/3) $\uparrow$}&\multicolumn{3}{c}{ECE (1/2/3) $\downarrow$}\\ \midrule
\multirow{3}{*}{ResNet-50} &SGD & 76.97 & 3.39 & 1.80 & 3.31&94.01 & 36.89&  35.81&24.99&7.97 & 4.23&17.29\\
 &SAM &77.32 &1.52  & 1.54 & 1.44 & 94.35 & 37.45 &  36.35&27.85&4.91 & 3.74&6.92\\ 
 &CSAM &\textbf{77.95} &\textbf{1.18}  & \textbf{1.09} & \textbf{1.19} & \textbf{94.67} & \textbf{38.29} &  \textbf{37.11}&\textbf{28.69}&\textbf{3.28} & \textbf{3.02}&\textbf{5.47}\\ \midrule
\multirow{3}{*}{ViT-S/32} & AdamW & 65.03 & 9.11 & 2.63  & 9.11&88.63 & 33.53&  32.87&26.48&14.73 & 12.28&19.57\\
 &SAM  & 69.21& 3.04 & 1.18  & 3.05 & 91.01 & 37.95&   36.09&33.36&3.35 & 6.27&7.89\\
 &CSAM  & \textbf{70.01}& \textbf{2.88} & \textbf{0.92}  & \textbf{2.78} & \textbf{91.54} & \textbf{38.88}&   \textbf{36.94}&\textbf{34.16}&\textbf{3.01} & \textbf{5.76}&\textbf{5.41}\\ \midrule
\multirow{3}{*}{ViT-S/16} &AdamW & 71.35 & 9.72 & 3.66 & 9.72&90.61 & 37.40&  35.54&24.26&14.14 & 12.27&18.63\\
 & SAM & 75.42& 1.76 & 1.66 & 1.73 & 93.27 & 43.36 &   39.15 &28.93&2.92 & 3.58 &5.02\\
 & CSAM & \textbf{75.91}& \textbf{1.58} & \textbf{1.34} & \textbf{1.54} & \textbf{93.66} & \textbf{44.01} &   \textbf{39.82} &\textbf{29.57}&\textbf{2.81} & \textbf{3.24} &\textbf{4.75}\\ \bottomrule
\end{tabular}
}
\end{table}
\par
As shown in Table \ref{tab:imagenet result}, SAM and CSAM consistently improve the test accuracy on ImageNet-1K validation set, though being more pronounced for ViTs ($\sim$ 4\%).
Meanwhile, ViTs are generally less calibrated than ResNet, which is somewhat inconsistent with the findings of \citep{minderer2021revisiting}.
One explanation might be that their comparison is based on the pretrained neural networks rather than training them from scratch.
But when models are trained by SAM, both of them achieve a much lower calibration error, and their gap becomes negligible.
For ImageNet-C, we consider three kinds of corruption: 1--motion blur, 2--defocus blur, and 3--impulse noise.
For each kind of corruption, we further average the accuracy and ECE across the five different severities.
Consistent with previous findings, Table \ref{tab:imagenet result} also indicates that SAM generalizes better than SGD and that ViTs trained by AdamW also tend to be less calibrated on ImageNet-C.
Interestingly, however, we observe that while ViT-S/16-SAM generalizes and calibrates worse than ResNet-50-SAM, it performs much better than the latter.
This might arise from the different implicit biases of SGD and AdamW.
\subsection{CSAM even calibrates better than SAM}
\label{sec: CSAM even calibrates better than SAM}
In this section, we attempt to compare CSAM and SAM against other popular baselines, including: focal loss \citep{mukhoti2020calibrating} that implicitly penalizes the gradient norms of confident examples and its two variants---DualFocal \citep{tao2023dual} and AdaFocal \citep{NEURIPS2022_0a692a24}, mixup \citep{zhang2018mixup} that implicitly performs label smoothing \citep{carratino2022mixup} to avoid the overconfidence issue, MMCE \citep{kumar2018trainable} that acts as a continuous and differentiable calibration error regulariser, MIT-L \citep{wang2023pitfall} that involves mixup inference in training, BatchEnsemble \citep{wenbatchensemble2020}, ACLS \citep{park2023acls}, BalCAL \citep{ni2025balancing}, and several probabilistic approaches---Rank1-BNN \citep{dusenberry2020efficient}, VI \citep{ovadia2019can}, MIMO \citep{havasitraining2021}, and bSAM \citep{mollenhoffsam2023}.
The backbone is WideResNet-20-10 \citep{zagoruyko2016wide}, and we generally follow the recommended setting to reproduce the results of each baseline.
The perturbation radius $\rho$ of SAM and CSAM is 0.2 for CIFAR-10/100, and we vary the hyper-parameter $\gamma$ of CSAM in \{0.5, 1.0, 2.0\}.
\begin{table}[t]
	\caption{Performance comparison between different methods on CIFAR-10. The results are averaged over 3 random seeds, with standard deviation displayed as well.}
	\label{tab:large scale cifar-10}
\resizebox{\textwidth}{!}{
	\begin{tabular}{ccccccc}
		\toprule
		& \textbf{Test Acc} $\uparrow$ & \textbf{ECE} $\downarrow$ & \textbf{ClasswiseECE} $\downarrow$ & \textbf{AdaECE} $\downarrow$ & \textbf{TCE} $\downarrow$ & \textbf{AUROC} $\uparrow$\\ \midrule
		CE & 95.83 $\pm$ 0.21 & 2.36 $\pm$ 0.11 & 0.52 $\pm$ 0.01 & 2.04 $\pm$ 0.11 & 1.06 $\pm$ 0.19 & 98.68 $\pm$ 0.04 \\
		Focal Loss (FL) & 95.91 $\pm$ 0.02 & 1.16 $\pm$ 0.13 & 0.38$\pm$ 0.01 & 1.42 $\pm$ 0.09 & 1.01 $\pm$ 0.28 & 99.04 $\pm$ 0.01 \\
		DualFocal & 95.73 $\pm$ 0.10 & 1.74 $\pm$ 0.09& 0.48 $\pm$ 0.02& 1.64 $\pm$ 0.07& 1.00 $\pm$ 0.09 & 99.26 $\pm$ 0.02 \\
		AdaFocal & 95.78 $\pm$ 0.06 & 0.91 $\pm$ 0.14 & 0.35 $\pm$ 0.01 & 0.65 $\pm$ 0.04 & 0.97 $\pm$ 0.08 & 99.10 $\pm$ 0.04 \\
		Mixup & 96.34 $\pm$ 0.10& 2.21 $\pm$ 1.11& 0.45 $\pm$ 0.21& 1.63 $\pm$ 1.04& 1.33 $\pm$ 0.25 & 99.12 $\pm$ 0.02 \\
		MIT-L & 96.56 $\pm$ 0.16 & 1.05 $\pm$ 0.02 & 0.31 $\pm$ 0.01 & 1.05 $\pm$ 0.05 & 0.57 $\pm$ 0.11 & 99.12 $\pm$ 0.03 \\
		MMCE & 95.94 $\pm$ 0.02 & 2.47 $\pm$ 0.04 & 0.54 $\pm$ 0.02 & 2.42 $\pm$ 0.04 & 1.15 $\pm$ 0.18 & 98.65 $\pm$ 0.05 \\
		BatchEnsemble & 95.92 $\pm$ 0.11 & 1.91 $\pm$ 0.06 & 0.45 $\pm$ 0.01 & 1.85 $\pm$ 0.03 & 0.41 $\pm$ 0.01 & 98.96 $\pm$ 0.01 \\
		Rank1-BNN & 95.50 $\pm$ 0.14 & 1.92 $\pm$ 0.29 & 0.45 $\pm$ 0.06 & 1.94 $\pm$ 0.29 & 0.51 $\pm$ 0.03 & 98.81 $\pm$ 0.12 \\
		VI & 94.33 $\pm$ 0.10 & 3.14 $\pm$ 0.12 & 0.69 $\pm$ 0.03 & 3.06 $\pm$ 0.15 & 0.76 $\pm$ 0.08 & 98.28 $\pm$ 0.07 \\
		MIMO & 95.96 $\pm$ 0.06 & 0.88 $\pm$ 0.06 & 0.33 $\pm$ 0.01 & 0.73 $\pm$ 0.08 & 0.74 $\pm$ 0.20 & 99.16 $\pm$ 0.01 \\
        ACLS & 95.91 $\pm$ 0.08 & 2.48 $\pm$ 0.07 & 0.55 $\pm$ 0.01 & 2.45 $\pm$ 0.07 & 1.09 $\pm$ 0.08 & 98.62 $\pm$ 0.01 \\
        % CPC & 95.81 $\pm$ 0.07 & 2.72 $\pm$ 0.13 & 0.58 $\pm$ 0.02 & 2.67 $\pm$ 0.10 & 0.87 $\pm$ 0.05 & 98.38 $\pm$ 0.01 \\
        % MBLS & 95.68 $\pm$ 0.19 & 2.62 $\pm$ 0.07 & 0.57 $\pm$ 0.01 & 2.63 $\pm$ 0.06 & 0.96 $\pm$ 0.08 & 98.58 $\pm$ 0.03 \\
        % MDCA & 96.00 $\pm$ 0.05 & 2.43 $\pm$ 0.06 & 0.52 $\pm$ 0.01 & 2.40 $\pm$ 0.02 & 0.83 $\pm$ 0.04 & 98.67 $\pm$ 0.02 \\
        BalCAL & 96.23 $\pm$ 0.09 & 1.89 $\pm$ 0.07 & 0.42 $\pm$ 0.01 & 1.93 $\pm$ 0.01 & 0.79 $\pm$ 0.05 & 98.77 $\pm$ 0.44 \\
        \midrule
		bSAM & 96.45 $\pm$ 0.03 & 1.82 $\pm$ 0.10 & 0.43 $\pm$ 0.02 & 1.78 $\pm$ 0.10 & 0.70 $\pm$ 0.23 & 98.95 $\pm$ 0.06 \\
		SAM & 96.91 $\pm$ 0.14 & 0.86 $\pm$ 0.13 & 0.26 $\pm$ 0.02 & 0.84 $\pm$ 0.14 & 0.52 $\pm$ 0.09 & 99.30 $\pm$ 0.02 \\
        % \rowcolor[HTML]{EBF4FF}
		CSAM & \textbf{96.97 $\pm$ 0.05} & \textbf{0.50 $\pm$ 0.03} & \textbf{0.23 $\pm$ 0.01} & \textbf{0.48 $\pm$ 0.03} & \textbf{0.47 $\pm$ 0.05} & \textbf{99.53 $\pm$ 0.02} \\ \bottomrule
	\end{tabular}
}
\end{table}
\par
From Tables \ref{tab:large scale cifar-10} and \ref{tab:large scale cifar-100}, we can observe that while focal loss generally hurts generalization, it does reduce the calibration error.
This observation also applies to the probabilistic approaches, such as Rank1-BNN and MIMO.
As a comparison, SAM significantly reduces the calibration error and is competitive, even superior to other baselines in many cases.
Note that the Bayesian variant, bSAM, does not perform better than SAM. The reason might be that it additionally introduces several hyperparameters, making it more difficult to tune and apply.
In contrast, the proposed CSAM further decreases the calibration error while simultaneously achieving a competitive generalization performance to SAM.
And when compared to other baselines, CSAM always achieves the lowest error, showing its versatility in generalization and calibration.
While our current study is limited to the cross-entropy loss, preliminary studies (see Table \ref{tab: sam with different losses}) indicate that SAM/CSAM can be further integrated with other training losses.
More results, such as sensitivity analysis of the hyperparameters and comparison to other variants of SAM, can be found in Appendices \ref{sec:Effects of Perturbation Radius} and \ref{sec: More Experimental Results on CSAM}.
\section{Conclusion}
\label{sec: conclusion}
Besides its well-known generalization benefits, we showed that SAM also excels at calibrating deep neural networks.
We proved that SAM achieves this goal by imposing an implicit regularization on the negative entropy of the predictive distribution during training (see Equation \ref{eq:theorem one sam}), which is similar to focal loss \citep{mukhoti2020calibrating}.
We further proposed a variant of SAM to improve calibration and validated its superiority across a number of networks and datasets.
% Moreover, studying when the calibration takes place during training (see Appendix \ref{sec:Calibration Performance of SAM at Different Stages of Training}) can also help us to design more computationally efficient optimizers.
% finally conducted extensive experiments to confirm the calibration superiority of both methods.
% \section*{Ethics statement}
% The authors read the ICLR Code of Ethics and declare no competing interests. 
\section*{Acknowledgments}
This work was supported in part by the National Natural Science Foundation
of China under Grants 12501710, 12301656, 62276208, 12326607 and 12371512, in part by the Natural Science Basic Research Program of Shaanxi Province under Grant 2024JC-JCQN-02, and in part by Fundamental Research Funds for the Central Universities.
\section*{Disclosure of the Use of Large Language Models}
The authors use large language models to assist with copyediting and to improve the overall readability of the text. The authors review and edit the manuscript and assume full responsibility for its final content. 
\bibliography{main}
\bibliographystyle{iclr2026_conference}

% \newpage
% \counterwithin*{figure}{section} % 
% \counterwithin*{table}{section}  %
% \renewcommand{\figurename}{Fig.}
\renewcommand{\thefigure}{S\arabic{figure}}
\renewcommand{\thetable}{S\arabic{table}}
\appendix
\onecolumn
\section{Theoretical Proofs}
\label{appendix: missing proofs}
\begin{figure}[t]
\centering
\begin{subfigure}[b]{0.31\linewidth}
    \centering
    \includegraphics[width=\textwidth, clip, trim= 0 0 0 0]{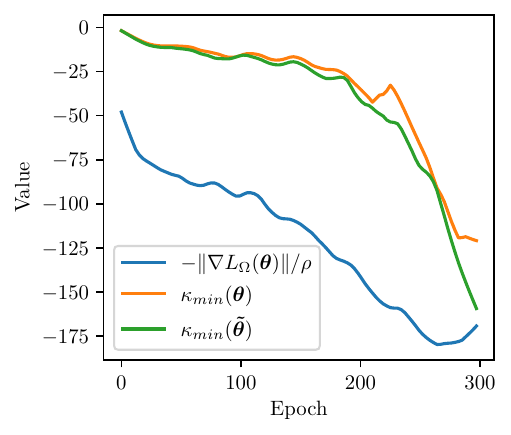}
    \caption{}
\end{subfigure}
\begin{subfigure}[b]{0.31\linewidth}
    \centering
    \includegraphics[width=\textwidth, clip, trim= 0 0 0 0]{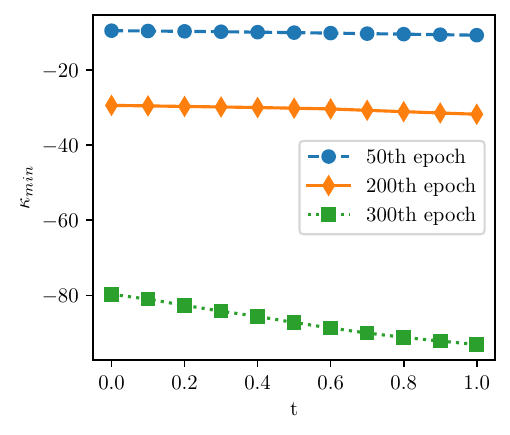}
    \caption{}
\end{subfigure}
\begin{subfigure}[b]{0.31\linewidth}
    \centering
    \includegraphics[width=\textwidth, clip, trim= 0 0 0 0]{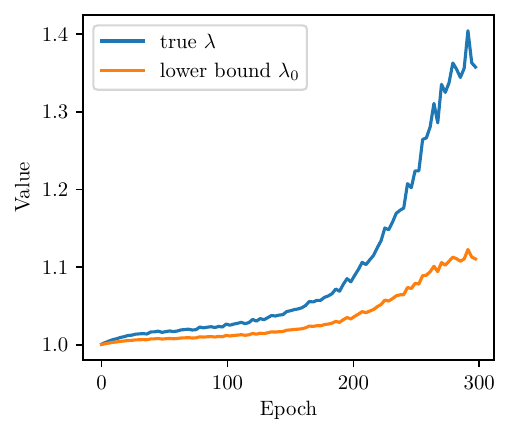}
    \caption{}
\end{subfigure}
\caption{To verify whether the boundedness assumption of $\kappa_{min}$ (see Lemmas \ref{lemma: one-sam} and \ref{lemma: m-sam}) holds for realistic neural networks, we further trained a ViT on the realistic ImageNet-1K dataset. (a) compares the value of $\kappa_{min}$ at the two endpoints $\boldsymbol{\theta}$ and $\tilde{\boldsymbol{\theta}}$ throughout the training process. (b) further illustrates how $\kappa_{min}$ evolves along the path from $\boldsymbol{\theta} (t=0)$ to $\tilde{\boldsymbol{\theta}} (t=1)$ at 50th, 200th, and 300th epoch, respectively. (c) records how the true coefficient $\lambda$ and its lower bound $\lambda_0$ (see Equation \ref{eq: ratio lower bound}) vary during training.}
\label{fig: vit monotone function}
\end{figure}

In this section, we first present the pseudocode of CSAM (Algorithm \ref{algo: csam}) and then the missing proofs in Section \ref{sec: theoretical analysis}.
We also validate the boundedness assumption on the ImageNet-1K dataset, as shown in Figure \ref{fig: vit monotone function}.
\begin{algorithm}[t]
	\caption{CSAM Optimizer}
	\begin{algorithmic}[1]
		\renewcommand{\algorithmicrequire}{ \textbf{Input:}}
		\REQUIRE Training set $S=\{z_i=(x_i, y_i)\}_{i=1}^n$, objective function $L_S(\boldsymbol{\theta})$, initial weight $\boldsymbol{\theta}_0\in\mathbb{R}^d$, learning rate $\eta>0$, perturbation radius $\rho>0$, training iterations $T$, regularization coefficient $\gamma>0$, and base optimizer $\mathcal{A}$ (e.g. SGD)
		\renewcommand{\algorithmicensure}{ \textbf{Output:}}
		\ENSURE $\boldsymbol{\theta}_T$
		\FOR{$t=0, 1, \cdots, T-1$}
		\STATE Sample a mini-batch $\Omega_t=\{z_1^t, \cdots, z_m^t\}$;
		\STATE Compute cross-entropy loss $L_{\Omega_t}(\boldsymbol{\theta_t}) = \frac{1}{m} \sum_{z_i\in \Omega_t} \ell_{\tilde{\boldsymbol{\theta}}} (z_i)$;
		\STATE Compute perturbed weight $\tilde{\boldsymbol{\theta}}_t = \boldsymbol{\theta}_t + \rho \cdot \frac{\nabla_{\boldsymbol{\theta}_t} L_{\Omega_t}(\boldsymbol{\theta}_t)}{ \|\nabla_{\boldsymbol{\theta}_t} L_{\Omega_t}(\boldsymbol{\theta}_t)\|}$;
		\STATE Compute perturbed loss $L_{\Omega_t}(\tilde{\boldsymbol{\theta}}_t)= \frac{1}{m} \sum_{z_i\in \Omega_t} \textcolor{red}{\tilde{\ell}_{\tilde{\boldsymbol{\theta}}} (z_i)}$ per Equation \eqref{eq: refined loss function};
		\STATE Compute gradient $\tilde{\boldsymbol{g}}_t=\nabla_{\boldsymbol{\theta}} L_{\Omega_t}(\tilde{\boldsymbol{\theta}}_t)|_{\boldsymbol{\theta}=\tilde{\boldsymbol{\theta}}_t}$ of the loss over the same $\Omega_t$;
		\STATE Update weight with base optimizer $\mathcal{A}$, e.g. $\boldsymbol{\theta}_{t+1} = \boldsymbol{\theta}_t - \eta \tilde{\boldsymbol{g}}_t$;
		\ENDFOR
	\end{algorithmic}
	\label{algo: csam}
\end{algorithm}
% \textbf{\emph{Do not put content after the references.}}
% %
% Put anything that you might normally include after the references in a separate
% supplementary file.

% We recommend that you build supplementary material in a separate document.
% If you must create one PDF and cut it up, please be careful to use a tool that
% doesn't alter the margins, and that doesn't aggressively rewrite the PDF file.
% pdftk usually works fine. 

% \textbf{Please do not use Apple's preview to cut off supplementary material.} In
% previous years it has altered margins, and created headaches at the camera-ready
% stage. 
% \subsection{Proof of Lemma \ref{lemma: one-sam}}
\paragraph{Proof of Lemma \ref{lemma: one-sam}.} According to the Taylor theorem, there always exists some $\boldsymbol{\theta}^\prime$ such that
\begin{align*}
    -\log \tilde{\mathbf{p}}_{y} = \ell_{\tilde{\boldsymbol{\theta}}}(z) & = \ell_{\boldsymbol{\theta}}(z) + (\tilde{\boldsymbol{\theta}} - \boldsymbol{\theta})^T\nabla \ell_{\boldsymbol{\theta}}(z) + \frac{1}{2}(\tilde{\boldsymbol{\theta}} - \boldsymbol{\theta})^T \nabla^2 \ell_{\boldsymbol{\theta}^\prime}(z)(\tilde{\boldsymbol{\theta}} - \boldsymbol{\theta}).
\end{align*}
Since $\tilde{\boldsymbol{\theta}} = \boldsymbol{\theta} + \rho \nabla \ell_{\boldsymbol{\theta}}(z)/\|\nabla \ell_{\boldsymbol{\theta}}(z)\|_2$ and $\kappa_{min}(\nabla^2 \ell_{\boldsymbol{\theta}^\prime}(z))>-\|\nabla \ell_{\boldsymbol{\theta}}(z)\|_2/\rho$, it follows that
\begin{align*}
    -\log \tilde{\mathbf{p}}_{y} = \ell_{\tilde{\boldsymbol{\theta}}}(z) & \geq \ell_{\boldsymbol{\theta}}(z) + \rho\|\nabla \ell_{\boldsymbol{\theta}}(z)\|_2 + \frac{\rho^2}{2}\kappa_{min}(\nabla^2 \ell_{\boldsymbol{\theta}^\prime}(z)) \geq -\log \mathbf{p}_{y} + \frac{\rho}{2}\|\nabla \ell_{\boldsymbol{\theta}}(z)\|_2,
\end{align*}
thus concluding the proof.
% \subsection{Proof of Theorem \ref{theorem: one-sam}}
\paragraph{Proof of Theorem \ref{theorem: one-sam}.}
It follows from Lemma \ref{lemma: one-sam} that $\tilde{\mathbf{p}}_{y} \leq \mathbf{p}_{y}$.
Recall that
\begin{align*}
    \ell_{\tilde{\boldsymbol{\theta}}}(z) = -\log \tilde{\mathbf{p}}_{y} 
     = \ell_{\boldsymbol{\theta}}(z) + \log \frac{\mathbf{p}_y}{\tilde{\mathbf{p}}_y} 
    & \geq \ell_{\boldsymbol{\theta}}(z) + \tilde{\mathbf{p}}_y \log \frac{\mathbf{p}_y}{\tilde{\mathbf{p}}_y} + (1-\tilde{\mathbf{p}}_y) \log \frac{1-\mathbf{p}_y}{1-\tilde{\mathbf{p}}_y} \\
    & \geq \ell_{\boldsymbol{\theta}}(z) - \frac{1-\tilde{\mathbf{p}}_y}{1-\mathbf{p}_y} H(\mathbf{p}_y) + H(\tilde{\mathbf{p}}_y),
\end{align*}
thus concluding the proof.
% \subsection{Proof of Lemma \ref{lemma: m-sam}}
\paragraph{Proof of Lemma \ref{lemma: m-sam}.}
There always exists some $\boldsymbol{\theta}^\prime \in \mathbb{R}^d$ such that
\begin{align*}
    -\log \left(\prod_{i=1}^m\tilde{\mathbf{p}}_{y_i}\right)^{1/m} &= -\frac{1}{m} \sum_{i=1}^m\log \tilde{\mathbf{p}}_{y_i} \\
    & = L_{\Omega}(\tilde{\boldsymbol{\theta}}) \\
    & = L_{\Omega}(\boldsymbol{\theta}) + \left(\tilde{\boldsymbol{\theta}} - \boldsymbol{\theta}\right)^T \nabla L_{\Omega}(\boldsymbol{\theta}) + \frac{1}{2}\left(\tilde{\boldsymbol{\theta}} - \boldsymbol{\theta}\right)^T \nabla^2 L_{\Omega}(\boldsymbol{\theta}^\prime)\left(\tilde{\boldsymbol{\theta}} - \boldsymbol{\theta}\right).
\end{align*}
A similar argument as Lemma \ref{lemma: one-sam} concludes the proof.
% \subsection{Proof of Theorem \ref{theorem: m-sam}}
\paragraph{Proof of Theorem \ref{theorem: m-sam}.}
The proof is straightforward.
According to the definition of $\mathbf{p}_y$ and $\tilde{\mathbf{p}}_y$, it yields that
\begin{align*}
    L_{\Omega}(\tilde{\boldsymbol{\theta}}) = -\log \tilde{\mathbf{p}}_{y} 
     = L_{\Omega}(\boldsymbol{\theta}) + \log \frac{\mathbf{p}_y}{\tilde{\mathbf{p}}_y} 
    & \geq L_{\Omega}(\boldsymbol{\theta}) + \tilde{\mathbf{p}}_y \log \frac{\mathbf{p}_y}{\tilde{\mathbf{p}}_y} + (1-\tilde{\mathbf{p}}_y) \log \frac{1-\mathbf{p}_y}{1-\tilde{\mathbf{p}}_y} \\
    & \geq L_{\Omega}(\boldsymbol{\theta}_k) - \frac{1-\tilde{\mathbf{p}}_y}{1-\mathbf{p}_y} H(\mathbf{p}_y) + H(\tilde{\mathbf{p}}_y),
\end{align*}
thus completing the proof.
% \subsection{Proof of Theorem \ref{theorem: csam}}
\paragraph{Proof of Theorem \ref{theorem: csam}.}
Recall that
\begin{align*}
	\tilde{\ell}_{\tilde{\boldsymbol{\theta}}}(z) = - (1 + \tilde{\mathbf{p}}_{y} )^{-\gamma}\log \tilde{\mathbf{p}}_{y} &\geq - (1 - \gamma \tilde{\mathbf{p}}_{y}) \log \tilde{\mathbf{p}}_{y}\\
	& = \ell_{\tilde{\boldsymbol{\theta}}}(z) + \gamma\tilde{\mathbf{p}}_y\log\tilde{\mathbf{p}}_y \\
	& \geq \ell_{\boldsymbol{\theta}}(z) - \frac{1-\tilde{\mathbf{p}}_y}{1-\mathbf{p}_y} H(\mathbf{p}_y) + H(\tilde{\mathbf{p}}_y) + \gamma\tilde{\mathbf{p}}_y\log\tilde{\mathbf{p}}_y \\
	& \geq \ell_{\boldsymbol{\theta}}(z) - \frac{1-\tilde{\mathbf{p}}_y}{1-\mathbf{p}_y} H(\mathbf{p}_y) + H(\tilde{\mathbf{p}}_y) + \frac{\gamma}{2}\tilde{\mathbf{p}}_y\log\tilde{\mathbf{p}}_y + \frac{\gamma}{2}(1 - \tilde{\mathbf{p}}_y) \log (1 - \tilde{\mathbf{p}}_y) \\
	& = \ell_{\boldsymbol{\theta}}(z) - \frac{1-\tilde{\mathbf{p}}_y}{1-\mathbf{p}_y} H(\mathbf{p}_y) + (1-\frac{\gamma}{2})H(\tilde{\mathbf{p}}_y),
\end{align*}
thus concluding the proof.
\section{Effects of Perturbation Radius $\rho$ and Coefficient $\gamma$}
\label{sec:Effects of Perturbation Radius}
\begin{figure}[h]
\centering
\includegraphics[width=0.65\linewidth, clip, trim= 0 0 0 0]{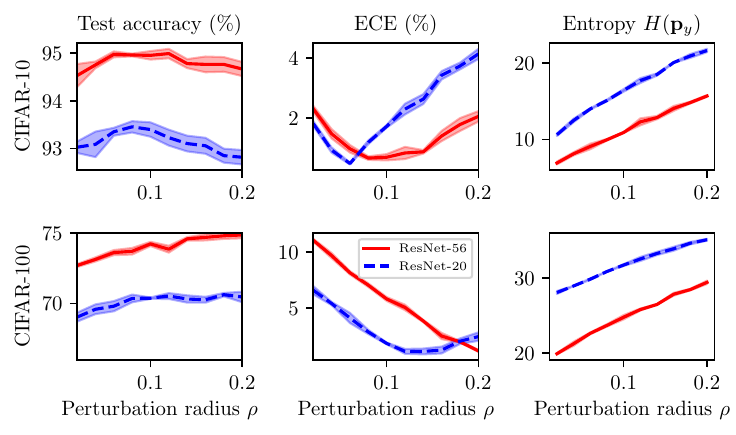}
\caption{Variation of different metrics for models trained under monotonically increasing perturbation radius $\rho$.
Note that $\mathbf{p}_y$ indicates the predicted probability associated with the true label in one-hot encoding, and $H(\mathbf{p}_y)$ is the corresponding entropy.
}
\label{fig: perturbation_radius_cifar10_ece}
\end{figure}

The perturbation radius $\rho$ is an important factor in determining the generalization performance \citep{foretsharpness2021}, but its effect on model calibration remains unknown.
To answer this question, we conduct another set of experiments while varying the perturbation radius $\rho$ from $0.02$ to $0.2$, an interval in which the optimal value of $\rho$ is often found.
Figure \ref{fig: perturbation_radius_cifar10_ece} shows that the entropy of the predictive distribution $H(\mathbf{p}_y)$ continues to increase for both models and datasets as expected.
However, we also observe that for both models the test accuracy on CIFAR-10 first increases and then decreases with the perturbation radius $\rho$, though the test accuracy on CIFAR-100 keeps increasing in this interval.
This implies that larger values of $\rho$ do not assure a better generalization.
On the other hand, the ECE on CIFAR-10 first decreases and then increases with the perturbation radius.
Moreover, the ECE of ResNet-56 is higher than that of ResNet-20 in the descending regime, which is aligned with the previous finding that increasing capacity by width or depth may hurt model calibration \citep{guo2017calibration}.
Meanwhile, when the perturbation radius exceeds the changing point, the ECE of ResNet-20 undergoes a sudden rise and becomes higher than that of ResNet-56, a phenomenon that is more pronounced for CIFAR-10 in this interval.
One explanation for this observation might be that models with low capacity are more amenable to the implicit regularization imposed by SAM.
The key point is that the perturbation radius $\rho$ should be relatively small to simultaneously achieve a lower ECE and a higher test accuracy than SGD.
\par
\begin{figure}[h]
\centering
\includegraphics[width=0.8\textwidth, clip, trim= 0 0 0 0]{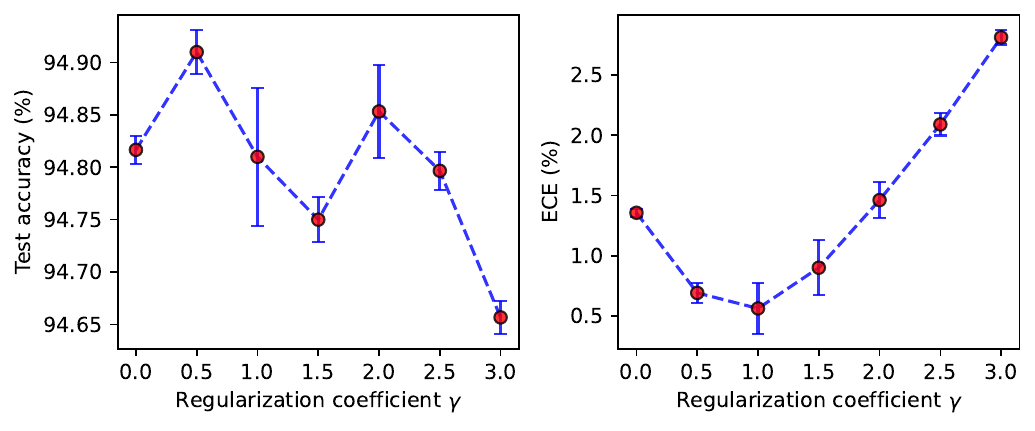}
\caption{Effects of CSAM hyperparameter $\gamma$ on test accuracy and ECE.}
\label{fig: gamma trend}
\end{figure}

And below we present how the additional hyperparameter $\gamma$ of CSAM affects the final generalization and calibration.
The base network is ResNet-56 trained on CIFAR-10, and the perturbation radius $\rho$ is 0.05.
We sweep $\gamma$ over $\{0, 0.5, 1.0, 1.5, 2.0, 2.5, 3.0\}$ and when $\gamma=0$, CSAM degenerates to the standard SAM.
As shown in Figure \ref{fig: gamma trend}, we can observe that when $\gamma=0.5$, CSAM improves both the generalization and calibration.
And the lowest value of ECE is attained when $\gamma=1$, but the test accuracy slightly decreases.
In contrast, increasing $\gamma$ up to 2 significantly deteriorates the performance.
Therefore, a relatively smaller value of $\gamma$ is preferred.
\par
Note that the perturbation radius $\rho$ has an important impact on $\lambda$ that controls the weight of the entropy term $-H(p_y)$. To investigate the interaction between $\rho$ and the hyper-parameter $\gamma$, we trained a number of ResNet-56 models on CIFAR-10/100 using different choices of $\rho$ and $\gamma$. Namely, $\rho$ from \{0.05, 0.1\} and $\gamma$ from \{0.0, 0.5, 1.0, 1.5, 2.0\}. Note that when $\gamma=0.0$, CSAM reduces to the standard SAM optimizer. From Table \ref{tab: interaction between rho and gamma}, we observe that CSAM can always generalize and calibrate better than SAM, when $\rho$ and $\gamma$ are carefully tuned.
Moreover, when training with a small value of $\rho$, it is suggested to combine with a relatively large value of $\gamma$, and vice versa. This is because when $\rho$ is large, the penalty coefficient $\lambda$ in Theorem \ref{theorem: one-sam} is also very large. Using a large $\gamma$ in this case will over-penalize the examples, which, as a result, adversely affects the generalization and calibration. 
\begin{table}[h]
\caption{A number of ResNet-56 were trained on CIFAR-10/100 datasets with varying $\rho$ and $\gamma$.}
\label{tab: interaction between rho and gamma}
\resizebox{\textwidth}{!}{
\begin{tabular}{llllllll}
\toprule
 & &  & $\gamma=0.0$ & $\gamma=0.5$ & $\gamma=1.0$ & $\gamma=1.5$ & $\gamma=2.0$ \\ \midrule
\multirow{4}{*}{CIFAR-10} & \multirow{2}{*}{Test Acc} & $\rho=0.05$ & 94.67 $\pm$ 0.06 & 94.61 $\pm$ 0.09 & 94.62 $\pm$ 0.14 & 94.67 $\pm$ 0.14 & \textbf{94.74 $\pm$ 0.06} \\ 
 & & $\rho=0.1$ & 94.74 $\pm$ 0.05 & 94.71 $\pm$ 0.07 & \textbf{94.90 $\pm$ 0.17} & 94.66 $\pm$ 0.09 & 94.79 $\pm$ 0.05 \\ \cmidrule{3-8}
 % & $\rho=0.2$ & 94.82 $\pm$ 0.16 & \textbf{94.85 $\pm$ 0.09} & 94.57 $\pm$ 0.22 & 94.64 $\pm$ 0.18 & 94.45 $\pm$ 0.08 \\ \midrule
& \multirow{2}{*}{ECE} & $\rho=0.05$ & 1.77 $\pm$ 0.04 & 1.46 $\pm$ 0.14  & 1.09 $\pm$ 0.07 & 0.66 $\pm$ 0.06 & \textbf{0.65 $\pm$ 0.14} \\ 
 & & $\rho=0.1$ & 0.82 $\pm$ 0.12 & \textbf{0.54 $\pm$ 0.11} & 0.69 $\pm$ 0.14 & 1.34 $\pm$ 0.33 & 2.04 $\pm$ 0.17 \\ \midrule
 % & $\rho=0.2$ & \textbf{0.91 $\pm$ 0.04} & 2.07 $\pm$ 0.09 & 3.06 $\pm$ 0.19 & 3.97 $\pm$ 0.17 & 4.84 $\pm$ 0.29 \\ \midrule
\multirow{4}{*}{CIFAR-100} & \multirow{2}{*}{Test Acc} & $\rho=0.05$ & 73.52 $\pm$ 0.42    & 73.48 $\pm$ 0.31     & 73.55 $\pm$ 0.18     & 73.57 $\pm$ 0.46 & \textbf{73.67 $\pm$ 0.18} \\
              & & $\rho=0.1$  & 73.94 $\pm$ 0.09    & 74.16 $\pm$ 0.17     & 73.96 $\pm$ 0.22 & \textbf{74.12 $\pm$ 0.09} & 73.86$\pm$ 0.39      \\ \cmidrule{3-8}
              % & $\rho=0.2$  & 74.92 $\pm$ 0.25    & **74.98 $\pm$ 0.10** & 74.49 $\pm$ 0.15     & 74.47 $\pm$ 0.22 & 74.49 $\pm$ 0.32     \\ \midrule
& \multirow{2}{*}{ECE}      & $\rho=0.05$ & 8.83 $\pm$ 0.15     & 7.52$\pm$ 0.14       & 6.24 $\pm$ 0.28      & 4.86 $\pm$ 0.37  & \textbf{3.67 $\pm$ 0.12}  \\
              & & $\rho=0.1$  & 6.17 $\pm$ 0.18     & 4.34 $\pm$ 0.45  & 3.26 $\pm$ 0.40      & 1.84 $\pm$ 0.14  & \textbf{1.42 $\pm$ 0.45}      \\ \bottomrule
              % & $\rho=0.2$  & 1.15 $\pm$ 0.12 & 0.90 $\pm$ 0.09      & 1.60 $\pm$ 0.22      & 3.11 $\pm$ 0.30  & 3.80 $\pm$ 0.51  \\ \bottomrule
\end{tabular}
}
\end{table}
\section{More Experimental Results on CSAM}
\label{sec: More Experimental Results on CSAM}
In this section, we include several networks like ResNets \citep{he2016deep}, Wide ResNets \citep{zagoruyko2016wide}, and PyramidNets \citep{han2017deep} to classify CIFAR-10/100.
Furthermore, the classical ResNet-18 for ImageNet-1K is further adapted to classify Tiny-ImageNet.
The initial learning rate and the weight decay coefficient are swept over \{0.01, 0.05, 0.1\} and \{1.0e-4, 5.0e-4, 1.0e-3\}, respectively.
By default, we use a mini-batch size of 128.
The optimizer is SGD with momentum 0.9, and the learning rate is scheduled in a cosine decay \citep{loshchilov2016sgdr}.
Note that all experiments are run on a GPU cluster with 2 cards, and it require approximately 1500 GPU hours in total.
As shown from Table \ref{tab: test accuracy} to Table \ref{tab: auroc before calibration}, CSAM consistently performs better than SAM, and it surpasses other baselines as well.
The code to reproduce these results is available at \url{https://drive.google.com/drive/folders/1O6up8Q7sdqekErGPmetIuMfEhsPZo-Hc?usp=sharing}.
While other variants of SAM potentially altered the loss landscape, their effects on calibration are supposed to be similar to SAM. This could be seen from Tables \ref{tab: sam variants on cifar10} and \ref{tab: sam variants on cifar100}, which summarize the results on CIFAR-10/100. Both SAM variants, including ASAM \citep{kwon2021asam} and VASSO \citep{li2023enhancing}, can reduce the calibration error, though ASAM generalizes much worse than SAM.
\par
To further demonstrate the efficacy of SAM on calibration, we also compare the result against several calibration-oriented training losses such as Label Smoothing (LS) and its variants, e.g., MBLS \citep{liu2022devil} and ACLS \citep{park2023acls}.
Besides them, popular approaches such as CPC \citep{cheng2022calibrating}, MDCA \citep{hebbalaguppe2022stitch}, and CRL \citep{moon2020confidence} are also included.
As shown in Table \ref{tab: ls comparison}, both Label Smoothing and its variants can reduce the ECE.
But, unfortunately, they also hurt the generalization performance.
In contrast, SAM and CSAM not only reduce the ECE but also significantly improve the test accuracy.
Moreover, we evaluate these methods on ImageNet-1K as well. The base model is ViT-S-32 and all runs are trained only once due to limited time. Table \ref{tab: ls imagenet1k} suggests that SAM/CSAM perform much better than other methods, both in generalization accuracy and calibration ECE, in ID and OOD settings.
\par
We are also interested in whether CSAM results in a flatter minima.
For this purpose, a number of ResNet-56 models are trained on CIFAR-100 with different optimizers. Table \ref{tab: csam reduces sharpness} indicates that CSAM can reach a flatter minima without compromising the calibration performance. Moreover, we also examine it with Focal Loss and Label Smoothing, both of which can significantly reduce the ECE value. It is interesting to find that they both converge to a flatter minima as well. This indicates a positive correlation between calibration and sharpness. 
\begin{table}[h]
	\caption{Performance comparison between different methods on CIFAR-100. The results are averaged over 3 random seeds, with standard deviation displayed as well.}
    \label{tab:large scale cifar-100}
	\resizebox{\textwidth}{!}{
	\begin{tabular}{ccccccc}
		\toprule
		& \textbf{Test Acc} $\uparrow$ & \textbf{ECE} $\downarrow$ & \textbf{ClasswiseECE} $\downarrow$ & \textbf{AdaECE} $\downarrow$ & \textbf{TCE} $\downarrow$ & \textbf{AUROC} $\uparrow$\\ \midrule
		CE & 81.01$\pm$ 0.11& 3.95 $\pm$ 0.28 & 0.21 $\pm$ 0.01& 3.86 $\pm$ 0.22 & 3.38 $\pm$ 0.41& 93.93 $\pm$ 0.05 \\
		Focal Loss (FL) & 80.55 $\pm$ 0.17 & 2.84 $\pm$ 0.36 & 0.19 $\pm$ 0.01& 2.79 $\pm$ 0.45 & 2.75 $\pm$ 0.36& 94.43 $\pm$ 0.01 \\
		DualFocal & 80.74 $\pm$ 0.24 & 2.68 $\pm$ 0.51& 0.18 $\pm$ 0.01& 2.66 $\pm$ 0.51& 2.24 $\pm$ 0.29& 94.81 $\pm$ 0.17 \\
		AdaFocal & 80.70 $\pm$ 0.11 & 2.58 $\pm$ 0.31 & 0.19 $\pm$ 0.01 & 2.61 $\pm$ 0.37 & 2.31 $\pm$ 0.29& 93.75 $\pm$ 0.05 \\
		Mixup & 82.09 $\pm$ 0.26 & 4.28 $\pm$ 0.27 & 0.18 $\pm$ 0.02 & 4.24 $\pm$ 0.31 & 4.20 $\pm$ 0.63 & 94.35 $\pm$ 0.08 \\
		MIT-L & 81.29 $\pm$ 0.18 & 3.26 $\pm$ 0.18 & 0.18 $\pm$ 0.01 & 3.24 $\pm$ 0.19 & 3.09 $\pm$ 0.49& 94.76 $\pm$ 0.12 \\
		MMCE & 81.02 $\pm$ 0.05 & 4.02 $\pm$ 0.29 & 0.18 $\pm$ 0.01 & 3.96 $\pm$ 0.22 & 3.69 $\pm$ 0.38& 93.83 $\pm$ 0.07 \\
		BatchEnsemble & 79.93 $\pm$ 0.11 & 6.86 $\pm$ 0.21 & 0.21 $\pm$ 0.01 & 6.77 $\pm$ 0.27 & 2.49 $\pm$ 0.17 & 94.15 $\pm$ 0.02 \\
		Rank1-BNN & 80.21 $\pm$ 0.06 & 3.59 $\pm$ 0.01 & 0.19 $\pm$ 0.01 & 3.57 $\pm$ 0.08 & 2.42 $\pm$ 0.11 & 94.29 $\pm$ 0.06 \\
		VI & 76.30 $\pm$ 0.06 & 10.29 $\pm$ 0.11 & 0.27 $\pm$ 0.03 & 10.29 $\pm$ 0.11 & 2.08 $\pm$ 0.35 & 92.62 $\pm$ 0.08 \\
		MIMO & 80.75 $\pm$ 0.13 & 2.38 $\pm$ 0.06 & 0.17 $\pm$ 0.01 & 2.31 $\pm$ 0.04 & 2.04 $\pm$ 0.01 & 95.14 $\pm$ 0.04 \\ 
        ACLS & 80.49 $\pm$ 0.19 & 6.38 $\pm$ 0.29 & 0.19 $\pm$ 0.01 & 6.31 $\pm$ 0.33 & 2.90 $\pm$ 0.47 & 93.16 $\pm$ 0.14 \\
        % CPC & 80.37 $\pm$ 0.22 & 6.22 $\pm$ 0.17 & 0.20 $\pm$ 0.01 & 6.16 $\pm$ 0.20 & 5.48 $\pm$ 0.26 & 92.31 $\pm$ 0.15 \\
        % MBLS & 80.49 $\pm$ 0.09 & 5.95 $\pm$ 0.36 & 0.19 $\pm$ 0.01 & 5.82 $\pm$ 0.46 & 3.22 $\pm$ 0.19 & 93.13 $\pm$ 0.08 \\
        % MDCA & 80.92 $\pm$ 0.04 & 3.87 $\pm$ 0.20 & 0.18 $\pm$ 0.01 & 3.80 $\pm$ 0.21 & 3.42 $\pm$ 0.33 & 93.92 $\pm$ 0.07 \\
        BalCAL & 81.34 $\pm$ 0.02 & 5.69 $\pm$ 0.18 & 0.18 $\pm$ 0.01 & 5.66 $\pm$ 0.24 & 2.64 $\pm$ 0.21 & 93.49 $\pm$ 0.09 \\
        \midrule
		bSAM & 80.59 $\pm$ 0.07 & 8.27 $\pm$ 0.13 & 0.22 $\pm$ 0.01 & 8.27 $\pm$ 0.14 & 2.59 $\pm$ 0.17 & 94.01 $\pm$ 0.11 \\
		SAM & 82.93 $\pm$ 0.15 & 2.11 $\pm$ 0.17 & 0.17 $\pm$ 0.01 & 2.17 $\pm$ 0.21& 1.89 $\pm$ 0.11& 94.15 $\pm$ 0.06 \\
        % \rowcolor[HTML]{EBF4FF}
		CSAM & \textbf{83.07 $\pm$ 0.19} & \textbf{1.93 $\pm$ 0.15} & \textbf{0.15 $\pm$ 0.01} & \textbf{1.99 $\pm$ 0.05} & \textbf{1.54 $\pm$ 0.30} & \textbf{96.07 $\pm$ 0.03} \\ \bottomrule
	\end{tabular}
}
\end{table}
\par
Finally, apart from the cross-entropy loss, we are also wondering whether the calibration benefit of SAM is persistent across other training losses.
As shown in Table \ref{tab: sam with different losses}, we can observe that when integrated with Focal Loss (FL) and ACLS \citep{park2023acls}, SAM still calibrates better than the baseline, whereas CSAM achieves the lowest ECE/TCE.
\begin{table}[h]
	\centering
	\caption{Results (mean$\pm$std) of test accuracy (\%) over 3 random runs.
		Text marked as bold indicates the best result.
	}
	\label{tab: test accuracy}
	\resizebox{\textwidth}{!}{
		\begin{tabular}{llllllllll>{\columncolor[HTML]{EBF4FF}}l}
			\toprule
			{} & {} & {\textbf{CE}} & \textbf{FL}   & \textbf{DualFocal}&\textbf{AdaFocal}& \textbf{Mixup} & \textbf{MMCE} & \textbf{MIT-L} & \textbf{SAM}  &\textbf{CSAM}\\ \midrule
			& ResNet-56 & 94.01 $\pm$ 0.15 & 93.99 $\pm$ 0.04& 93.84 $\pm$ 0.22&93.87 $\pm$ 0.08& 94.42 $\pm$ 0.15 & 94.19 $\pm$ 0.22 & 94.68 $\pm$ 0.08& {94.92 $\pm$ 0.24}  &\textbf{95.00 $\pm$ 0.25}
\\
			& WRN-28-10 & 95.83 $\pm$ 0.21 & 95.91 $\pm$ 0.02& 95.73 $\pm$ 0.10&95.78 $\pm$ 0.06& 96.64 $\pm$ 0.10 & 95.94 $\pm$ 0.02 & 96.56 $\pm$ 0.16&\textbf{96.91$\pm$ 0.14}  &96.87 $\pm$ 0.05
\\
			\multirow{-3}{*}{CIFAR-10}& PyramidNet-110 & 96.07 $\pm$ 0.23 & 96.03 $\pm$ 0.06& 96.14 $\pm$ 0.04&96.00 $\pm$ 0.11& 96.77 $\pm$ 0.08 & 96.13 $\pm$ 0.08 & 96.78 $\pm$ 0.17&{97.14 $\pm$ 0.06}  &\textbf{97.26 $\pm$ 0.03}
\\ \midrule
			& ResNet-56 & 72.06 $\pm$ 0.13 & 71.96 $\pm$ 0.28& 71.43 $\pm$ 0.04&72.00 $\pm$ 0.08& 74.15 $\pm$ 0.29 & 72.17 $\pm$ 0.12 & 74.28 $\pm$ 0.42&{74.71 $\pm$ 0.30}  &\textbf{74.95 $\pm$ 0.32}
\\
			& WRN-28-10 & 81.04 $\pm$ 0.11 & 80.55 $\pm$ 0.17& 80.74 $\pm$ 0.24&80.70 $\pm$ 0.11& 82.09 $\pm$ 0.26 & 81.02 $\pm$ 0.05 & 81.29 $\pm$ 0.18 &{82.93 $\pm$ 0.15}  &\textbf{83.05 $\pm$ 0.19}
\\
			\multirow{-3}{*}{CIFAR-100}& PyramidNet-110 & 81.21 $\pm$ 0.52 & 81.53 $\pm$ 0.12& 81.76 $\pm$ 0.07&81.81 $\pm$ 0.38& 82.94 $\pm$ 0.29 & 81.36 $\pm$ 0.31 & 82.41 $\pm$ 0.02&{84.08 $\pm$ 0.29}  &\textbf{84.16 $\pm$ 0.15}
\\ \midrule
			Tiny-ImageNet & ResNet-18 & 51.96 $\pm$ 0.35& 52.61 $\pm$ 0.59& 53.02 $\pm$ 0.86&50.36 $\pm$ 0.69& 51.45 $\pm$ 0.70& 51.31 $\pm$ 0.79& 51.97 $\pm$ 0.24&{56.81 $\pm$ 0.31} &\textbf{57.13 $\pm$ 0.96}\\ \bottomrule
		\end{tabular}
	}
\end{table}
\begin{table}[h]
	\centering
	\caption{Results (mean$\pm$std) of ECE (\%) with $M=15$ over 3 random runs.
		Text marked as bold indicates the best result.
	}
	\label{tab: ece before calibration}
	\resizebox{\textwidth}{!}{
		\begin{tabular}{llllllllll>{\columncolor[HTML]{EBF4FF}}l}
			\toprule
			{} & {} & {\textbf{CE}} & \textbf{FL}   & \textbf{DualFocal}&\textbf{AdaFocal}& \textbf{Mixup} & \textbf{MMCE} & \textbf{MIT-L} & \textbf{SAM}  &\textbf{CSAM}\\ \midrule
			& ResNet-56 & 3.89 $\pm$ 0.16 & 1.81 $\pm$ 0.12
			& 2.50 $\pm$ 0.03&0.89 $\pm$ 0.12& 3.87 $\pm$ 0.09 & 3.61 $\pm$ 0.17 & 1.83 $\pm$ 0.18 &{0.64 $\pm$ 0.09}  &\textbf{0.58 $\pm$ 0.07}
\\
			& WRN-28-10 & 2.36 $\pm$ 0.11 & 1.16 $\pm$ 0.13
			& 4.74 $\pm$ 0.09&0.91 $\pm$ 0.14& 4.66 $\pm$ 1.11 & 2.47 $\pm$ 0.04 & 1.05 $\pm$ 0.02  &{0.86 $\pm$ 0.13}  &\textbf{0.50 $\pm$ 0.03}
\\
			\multirow{-3}{*}{CIFAR-10}& PyramidNet-110 & 2.54 $\pm$ 0.19 & 1.17 $\pm$ 0.15
			& 4.64 $\pm$ 0.05&0.96 $\pm$ 0.12& 2.23 $\pm$ 0.84 & 2.49 $\pm$ 0.12 & 1.22 $\pm$ 0.14&{0.74 $\pm$ 0.08}  &\textbf{0.32 $\pm$ 0.06}
\\ \midrule
			& ResNet-56 & 13.29 $\pm$ 0.15 & 8.25 $\pm$ 0.23
			& 4.93 $\pm$ 0.06&1.71 $\pm$ 0.09& 2.43 $\pm$ 0.32 & 13.49 $\pm$ 0.19 &5.11 $\pm$ 1.38 &{1.66 $\pm$ 0.16}  &\textbf{0.84 $\pm$ 0.15}
\\
			& WRN-28-10 & 3.95 $\pm$ 0.28 & 2.84 $\pm$ 0.36
			& 12.66 $\pm$ 0.51&2.58 $\pm$ 0.31& 4.28 $\pm$ 0.27 & 4.02 $\pm$ 0.29 & 3.26 $\pm$ 0.18 &{2.11 $\pm$ 0.17}  &\textbf{1.50 $\pm$ 0.07}
\\
			\multirow{-3}{*}{CIFAR-100}& PyramidNet-110 & 9.52 $\pm$ 0.64 & 4.26 $\pm$ 0.39
			& 10.58 $\pm$ 0.55&1.95 $\pm$ 0.11& 3.25 $\pm$ 1.19 & 9.24 $\pm$ 0.38 & 3.03 $\pm$ 0.38&{1.91 $\pm$ 0.14}  &\textbf{1.69 $\pm$ 0.04}
\\
			\midrule
			Tiny-ImageNet & ResNet-18 & 7.65 $\pm$ 2.21& 4.35 $\pm$ 0.64& 16.30 $\pm$ 0.53&11.71 $\pm$ 0.66& 10.81 $\pm$ 0.66& 9.34 $\pm$ 2.10& 4.09 $\pm$ 0.17& {3.46 $\pm$ 0.15} &\textbf{2.75 $\pm$ 0.47}\\
			\bottomrule
		\end{tabular}
	}
\end{table}
\begin{table}[h]
\centering
\caption{Results (mean$\pm$std) of Classwise ECE (\%) with $M=15$ over 3 random runs.
Text marked as bold indicates the best result.
}
\label{tab: class-ece before calibration}
\resizebox{\textwidth}{!}{
\begin{tabular}{llllllllll>{\columncolor[HTML]{EBF4FF}}l}
\toprule
{} & {} & {\textbf{CE}} & \textbf{FL}   & \textbf{DualFocal}&\textbf{AdaFocal}& \textbf{Mixup} & \textbf{MMCE} & \textbf{MIT-L} & \textbf{SAM}  &\textbf{CSAM}\\ \midrule
 & ResNet-56 & 0.80 $\pm$ 0.01& 0.47 $\pm$ 0.02& 0.67 $\pm$ 0.01&0.37 $\pm$ 0.03& 0.87 $\pm$ 0.06& 0.78 $\pm$ 0.03& 0.46 $\pm$ 0.02&{0.32 $\pm$ 0.01} &\textbf{0.29 $\pm$ 0.02}
\\
 & WRN-28-10 & 0.52 $\pm$ 0.01& 0.38 $\pm$ 0.01& 1.11 $\pm$ 0.02&0.35 $\pm$ 0.00& 1.05 $\pm$ 0.23& 0.54 $\pm$ 0.02& 0.31 $\pm$ 0.01&{0.26 $\pm$ 0.02} &\textbf{0.23 $\pm$ 0.01}
\\
\multirow{-3}{*}{CIFAR-10}& PyramidNet-110 & 0.56 $\pm$ 0.03& 0.36 $\pm$ 0.02& 1.03 $\pm$ 0.03&0.34 $\pm$ 0.01& 0.45 $\pm$ 0.02& 0.55 $\pm$ 0.02& 0.31 $\pm$ 0.01&{0.25 $\pm$ 0.01} &\textbf{0.20 $\pm$ 0.01}
\\ \midrule
 & ResNet-56 & 0.32 $\pm$ 0.01& 0.25 $\pm$ 0.00& 0.21 $\pm$ 0.01&0.19 $\pm$ 0.00& 0.19 $\pm$ 0.00& 0.33 $\pm$ 0.00&0.20 $\pm$ 0.01&\textbf{0.16 $\pm$ 0.00} &\textbf{0.16 $\pm$ 0.00}
\\
 & WRN-28-10 & 0.18 $\pm$ 0.01& 0.19 $\pm$ 0.00& 0.34 $\pm$ 0.01&0.19 $\pm$ 0.00& 0.18 $\pm$ 0.01& 0.18 $\pm$ 0.01& 0.18 $\pm$ 0.01&{0.17 $\pm$ 0.00} &\textbf{0.15 $\pm$ 0.01}
\\
\multirow{-3}{*}{CIFAR-100}& PyramidNet-110 & 0.23 $\pm$ 0.01& 0.17 $\pm$ 0.00& 0.30 $\pm$ 0.01&0.17 $\pm$ 0.00& 0.18 $\pm$ 0.03& 0.23 $\pm$ 0.01& 0.16 $\pm$ 0.00&{0.15 $\pm$ 0.00} &\textbf{0.14 $\pm$ 0.01}
\\
\midrule
Tiny-ImageNet & ResNet-18 & 0.21 $\pm$ 0.01& \textbf{0.19 $\pm$ 0.00}& 0.24 $\pm$ 0.01&0.23 $\pm$ 0.01& 0.21 $\pm$ 0.01& 0.21 $\pm$ 0.01& \textbf{0.19 $\pm$ 0.00}& \textbf{0.19 $\pm$ 0.00} &\textbf{0.19 $\pm$ 0.00}\\
\bottomrule
\end{tabular}
}
\end{table}
\begin{table}[h]
\centering
\caption{Results (mean$\pm$std) of Adaptive ECE (\%) with $M=15$ over 3 random runs.
Text marked as bold indicates the best result.
}
\label{tab: adaptive-ece before calibration}
\resizebox{\textwidth}{!}{
\begin{tabular}{llllllllll>{\columncolor[HTML]{EBF4FF}}l}
\toprule
{} & {} & {\textbf{CE}} & \textbf{FL}   & \textbf{DualFocal}&\textbf{AdaFocal}& \textbf{Mixup} & \textbf{MMCE} & \textbf{MIT-L} & \textbf{SAM}  &\textbf{CSAM}\\ \midrule
 & ResNet-56 & 3.71 $\pm$ 0.06& 2.13 $\pm$ 0.19& 2.19 $\pm$ 0.18&1.03 $\pm$ 0.14& 3.97 $\pm$ 0.08& 3.55 $\pm$ 0.15& 1.83 $\pm$ 0.14&{0.90 $\pm$ 0.14} &\textbf{0.51 $\pm$ 0.03}
\\
 & WRN-28-10 & 2.36 $\pm$ 0.11& 1.42 $\pm$ 0.09& 4.64 $\pm$ 0.07&{0.65 $\pm$ 0.04}& 4.63 $\pm$ 1.04& 2.42 $\pm$ 0.04& 1.05 $\pm$ 0.05&0.84 $\pm$ 0.14 &\textbf{0.48 $\pm$ 0.04}
\\
\multirow{-3}{*}{CIFAR-10}& PyramidNet-110 & 2.53 $\pm$ 0.19& 1.78 $\pm$ 0.07& 4.54 $\pm$ 0.02&0.88 $\pm$ 0.09& 2.69 $\pm$ 0.22& 2.49 $\pm$ 0.13& 1.19 $\pm$ 0.17&{0.70 $\pm$ 0.05} &\textbf{0.19 $\pm$ 0.02}
\\ \midrule
 & ResNet-56 & 13.36 $\pm$ 0.12& 8.23 $\pm$ 0.26& 4.91 $\pm$ 0.06&1.82 $\pm$ 0.21& 2.48 $\pm$ 0.23& 13.48 $\pm$ 0.21&5.09 $\pm$ 1.36&{1.02 $\pm$ 0.02} &\textbf{0.96 $\pm$ 0.14}
\\
 & WRN-28-10 & 3.86 $\pm$ 0.22& 2.79 $\pm$ 0.45& 12.66 $\pm$ 0.51&{2.61 $\pm$ 0.37}& 4.24 $\pm$ 0.31& 3.96 $\pm$ 0.22& 3.24 $\pm$ 0.19&4.67 $\pm$ 0.21
 &\textbf{1.50 $\pm$ 0.01}
\\
\multirow{-3}{*}{CIFAR-100}& PyramidNet-110 & 9.29 $\pm$ 0.54& 4.06 $\pm$ 0.49& 10.58 $\pm$ 0.55&1.76 $\pm$ 0.22& 3.25 $\pm$ 1.01& 9.18 $\pm$ 0.41& 2.99 $\pm$ 0.02&{1.65 $\pm$ 0.14} &\textbf{1.45 $\pm$ 0.04}
\\
\midrule
Tiny-ImageNet & ResNet-18 & 7.55 $\pm$ 2.27& 4.25 $\pm$ 0.56& 16.31 $\pm$ 0.54&11.71 $\pm$ 0.66& 10.79 $\pm$ 0.64& 9.19 $\pm$ 2.12& {3.33 $\pm$ 0.14}& 4.07 $\pm$ 0.19 &\textbf{2.65 $\pm$ 0.30}\\
\bottomrule
\end{tabular}
}
\end{table}
\begin{table}[h]
\centering
\caption{Results (mean$\pm$std) of AUROC (\%) over 3 random runs.
Text marked as bold indicates the best result.
}
\label{tab: auroc before calibration}
\resizebox{\textwidth}{!}{
\begin{tabular}{llllllllll>{\columncolor[HTML]{EBF4FF}}l}
\toprule
{} & {} & {\textbf{CE}} & \textbf{FL}   & \textbf{DualFocal}&\textbf{AdaFocal}& \textbf{Mixup} & \textbf{MMCE} & \textbf{MIT-L} & \textbf{SAM}  &\textbf{CSAM}\\ \midrule
 & ResNet-56 & 97.98 $\pm$ 0.03& 98.47 $\pm$ 0.05& 98.73 $\pm$ 0.09&98.78 $\pm$ 0.03& 98.59 $\pm$ 0.03& 98.04 $\pm$ 0.02& 98.72 $\pm$ 0.03&{99.07 $\pm$ 0.07} &\textbf{99.19 $\pm$ 0.02}
\\
 & WRN-28-10 & 98.68 $\pm$ 0.04& 99.04 $\pm$ 0.01& 99.26 $\pm$ 0.02&99.10 $\pm$ 0.04& 99.12 $\pm$ 0.02& 98.65 $\pm$ 0.05& 99.12 $\pm$ 0.03&{99.30 $\pm$ 0.02} &\textbf{99.40 $\pm$ 0.01}
\\
\multirow{-3}{*}{CIFAR-10} & PyramidNet-110 & 98.64 $\pm$ 0.04& 98.96 $\pm$ 0.04& 99.40 $\pm$ 0.04&99.16 $\pm$ 0.04& 99.00 $\pm$ 0.03& 98.66 $\pm$ 0.04& 99.20 $\pm$ 0.02&{99.41 $\pm$ 0.03} &\textbf{99.52 $\pm$ 0.02}
\\ \midrule
 & ResNet-56 & 91.06 $\pm$ 0.01& 92.32 $\pm$ 0.09& 92.69 $\pm$ 0.09&93.44 $\pm$ 0.07& 92.87 $\pm$ 0.09& 90.99 $\pm$ 0.04&93.32 $\pm$ 0.27&{94.35 $\pm$ 0.05} &\textbf{94.57 $\pm$ 0.07}
\\
 & WRN-28-10 & 93.93 $\pm$ 0.05& 94.43 $\pm$ 0.01& {94.81 $\pm$ 0.17}&93.75 $\pm$ 0.05& 94.35 $\pm$ 0.08& 93.83 $\pm$ 0.07& 94.76 $\pm$ 0.12&{94.15 $\pm$ 0.06}
 &\textbf{96.06 $\pm$ 0.03}
\\
\multirow{-3}{*}{CIFAR-100} & PyramidNet-110 & 93.46 $\pm$ 0.22& 94.29 $\pm$ 0.01& 95.16 $\pm$ 0.03&95.03 $\pm$ 0.09& 94.62 $\pm$ 0.04& 93.47 $\pm$ 0.13& 95.21 $\pm$ 0.08&{96.05 $\pm$ 0.07} &\textbf{96.14 $\pm$ 0.05}
\\
\midrule
Tiny-ImageNet & ResNet-18 & 82.62 $\pm$ 0.37& 83.31 $\pm$ 0.63& 81.08 $\pm$ 0.42&81.32 $\pm$ 0.65& 79.32 $\pm$ 0.53& 82.57 $\pm$ 0.06& 82.38 $\pm$ 0.13& {85.63 $\pm$ 0.31} &\textbf{85.69 $\pm$ 0.15}\\
\bottomrule
\end{tabular}
}
\end{table}
\begin{table}[h]
\caption{Calibration performance of different SAM variants on CIFAR-10.}
\label{tab: sam variants on cifar10}
\resizebox{\textwidth}{!}{
\begin{tabular}{lllllll}
\toprule
 & Test Acc & ECE & ClasswiseECE & AdaECE & TCE & AUROC \\ \midrule
 SGD & 94.26 $\pm$ 0.06 & 3.52 $\pm$ 0.12 & 0.78 $\pm$ 0.01 & 3.52 $\pm$ 0.11 & 1.03 $\pm$ 0.01 & 98.09 $\pm$ 0.08 \\
SAM & 94.50 $\pm$ 0.20 & 1.83 $\pm$ 0.08 & 0.44 $\pm$ 0.01 & 1.78 $\pm$ 0.03 & 0.70 $\pm$ 0.12 & 98.78 $\pm$ 0.08 \\
ASAM & 94.82 $\pm$ 0.11 & 2.03 $\pm$ 0.11 & 0.48 $\pm$ 0.01 & 2.01 $\pm$ 0.14 & 0.61 $\pm$ 0.05 & 98.71 $\pm$ 0.02 \\
VASSO & 94.70 $\pm$ 0.10 & 1.75 $\pm$ 0.04 & 0.42 $\pm$ 0.01 & 1.69 $\pm$ 0.05 & 0.71 $\pm$ 0.09 & 98.87 $\pm$ 0.06 \\ \midrule
CSAM & \textbf{94.58 $\pm$ 0.14} & \textbf{1.47 $\pm$ 0.17} & \textbf{0.41 $\pm$ 0.02} & \textbf{1.41 $\pm$ 0.19} & \textbf{0.72 $\pm$ 0.20} & \textbf{98.87 $\pm$ 0.06} \\ \bottomrule
\end{tabular}
}
\end{table}
\begin{table}[h]
\caption{Calibration performance of different SAM variants on CIFAR-100.}
\label{tab: sam variants on cifar100}
\resizebox{\textwidth}{!}{
\begin{tabular}{lllllll}
\toprule
 & Test Acc & ECE & ClasswiseECE & AdaECE & TCE & AUROC \\ \midrule
 SGD & 72.00 $\pm$ 0.15 & 13.14 $\pm$ 0.25 & 0.33 $\pm$ 0.01 & 13.13 $\pm$ 0.25 & 1.64 $\pm$ 0.09 & 91.12 $\pm$ 0.11 \\
SAM & 74.87 $\pm$ 0.21 & 1.59 $\pm$ 0.10 & 0.17 $\pm$ 0.01 & 1.39 $\pm$ 0.16 & 1.38 $\pm$ 0.11 & 94.40 $\pm$ 0.01 \\
ASAM & 74.11 $\pm$ 0.11 & 6.48 $\pm$ 0.26 & 0.22 $\pm$ 0.01 & 6.43 $\pm$ 0.25 & 1.31 $\pm$ 0.31 & 93.13 $\pm$ 0.09 \\
VASSO & 74.94 $\pm$ 0.53 & 1.55 $\pm$ 0.25 & 0.17 $\pm$ 0.01 & 1.55 $\pm$ 0.24 & 1.43 $\pm$ 0.23 & 94.31 $\pm$ 0.07 \\ \midrule
CSAM & \textbf{74.85 $\pm$ 0.20} & \textbf{1.31 $\pm$ 0.40} & \textbf{0.17 $\pm$ 0.01} & \textbf{1.22 $\pm$ 0.23} & \textbf{1.36 $\pm$ 0.44} & \textbf{94.55 $\pm$ 0.06} \\ \bottomrule
\end{tabular}
}
\end{table}
\begin{table}[h]
\caption{Comparison against other calibration-oriented training losses.}
\label{tab: ls comparison}
\resizebox{\textwidth}{!}{
\begin{tabular}{@{}llllllllll@{}}
\toprule
    & SGD              & LS         & ACLS             & CRL              & CPC              & MBLS             & MDCA             & SAM              & CSAM                 \\ \midrule
Test Acc & 72.30 $\pm$ 0.12 & 72.54 $\pm$ 0.18 & 72.49 $\pm$ 0.05 & 72.36 $\pm$ 0.18 & 72.45 $\pm$ 0.47 & 72.42 $\pm$ 0.24 & 72.38 $\pm$ 0.48 & 74.63 $\pm$ 0.24 & \textbf{74.98 $\pm$ 0.16} \\
ECE & 13.03 $\pm$ 0.26 & 4.38 $\pm$ 0.14  & 2.32 $\pm$ 0.40  & 3.43 $\pm$ 0.25  & 2.62 $\pm$ 0.54  & 2.33 $\pm$ 0.42  & 3.16 $\pm$ 0.17  & 1.87 $\pm$ 0.25  & \textbf{1.25 $\pm$ 0.15} \\ \bottomrule
\end{tabular}
}
\end{table}

\begin{table}[h]
\centering
\caption{Results on the ImageNet-1K dataset. Slightly different from the custom setting, we reserve 20\% of the ImageNet-1K validation set as a new validation set for early stopping and temperature scaling, and the remaining images therefore constitute a test set. Both metrics are evaluated on the test set (1/2/3 indicate different types of corruption).}
\label{tab: ls imagenet1k}
% \resizebox{\textwidth}{!}{
\begin{tabular}{lcccccccc}
\toprule
     &            \multicolumn{2}{c}{\textbf{ID Metrics}}                &  \multicolumn{6}{c}{\textbf{OOD Metrics}}        \\ \midrule
     & Test Acc $\uparrow$ & ECE $\downarrow$ &           \multicolumn{3}{c}{Test Acc (1/2/3) $\uparrow$}   &  \multicolumn{3}{c}{ ECE (1/2/3) $\downarrow$ }          \\ \midrule
SGD  & 65.91               & 11.56            & 33.54     & 30.21                       & 34.18       & 23.70    & 24.65                    & 18.00    \\
MBLS & 67.19               & 2.21             & 34.54     & 31.58                       & 35.58       & 11.83    & 12.61                    & 8.58     \\
CPC  & 65.68               & 9.34            & 32.34     & 31.01                       & 35.71       & 20.46    & 23.19                    & 15.04    \\
ACLS & 66.68               & 3.97             & 34.67     & 31.88                       & 36.00       & 15.72    & 16.17                    & 10.25    \\
MDCA & 65.23               & 8.38            & 33.11      & 29.91                       & 33.71       & 23.31    & 24.28                    & 17.15    \\
SAM  & 69.48               & 2.47             & 38.25     & 35.27                       & 39.54       & 7.33     & 7.77                     & 3.15     \\
CSAM & \textbf{69.78}          & \textbf{1.79}         & \textbf{38.58} & \textbf{35.46}                  & \textbf{39.85}   & \textbf{7.07} & \textbf{7.70}                 & \textbf{2.76} \\ \bottomrule
\end{tabular}
% }
\end{table}

\begin{table}[h]
\caption{Performance of CSAM/SAM when combined with other training losses (e.g. FL and ACLS).}
\label{tab: sam with different losses}
\resizebox{\textwidth}{!}{
\begin{tabular}{@{}lcccccc@{}}
\toprule
          & \multicolumn{3}{c}{ResNet-56}                             & \multicolumn{3}{c}{WRN-28-10}                             \\ \midrule
          & Test Acc $\uparrow$ & ECE $\downarrow$ & TCE $\downarrow$ & Test Acc $\uparrow$ & ECE $\downarrow$ & TCE $\downarrow$ \\ \midrule
FL        & $71.96 \pm 0.28$    & $8.25 \pm 0.23$  & $3.27 \pm 0.16$  & $80.55 \pm 0.17$    & $2.84 \pm 0.36$  & $2.75 \pm 0.36$  \\
FL+SAM    & $73.11 \pm 0.01$    & $6.90 \pm 0.38$  & $1.90 \pm 0.13$  & $81.33 \pm 0.23$    & $2.41 \pm 0.06$  & $1.76 \pm 0.12$  \\
FL+CSAM   & \textbf{73.51 $\pm$ 0.10}    & \textbf{3.94 $\pm$ 0.17}  & \textbf{1.58 $\pm$ 0.13}  & \textbf{82.08 $\pm$ 0.08}    & \textbf{1.88 $\pm$ 0.11}  & \textbf{1.42 $\pm$ 0.16}  \\ \midrule
ACLS      & $72.55 \pm 0.08$    & $5.88 \pm 0.18$  & $3.29 \pm 0.25$  & $80.49 \pm 0.19$    & $6.38 \pm 0.29$  & $2.90 \pm 0.47$  \\
ACLS+SAM  & $74.53 \pm 0.09$    & $1.37 \pm 0.05$  & $1.27 \pm 0.18$  & $83.04 \pm 0.01$    & $1.87 \pm 0.10$  & $1.80 \pm 0.12$  \\
ACLS+CSAM & \textbf{74.86 $\pm$ 0.07}    & \textbf{1.01 $\pm$ 0.12}  & \textbf{0.85 $\pm$ 0.05}  & \textbf{83.13 $\pm$ 0.08}    & \textbf{1.37 $\pm$ 0.21}  & \textbf{1.36 $\pm$ 0.06} \\ \bottomrule
\end{tabular}
}
\end{table}

\begin{table}[h]
\caption{Comparison of sharpness (measured by the largest eigenvalue $\kappa_{max}$ of the Hessian) of different optimizers.}
\label{tab: csam reduces sharpness}
\resizebox{\textwidth}{!}{
\begin{tabular}{@{}llllll@{}}
\toprule
                                        & SGD                & Focal Loss         & Label Smoothing    & SAM                & CSAM                   \\ \midrule
ECE                        & 13.03 $\pm$ 0.26   & 1.85 $\pm$ 0.09    & 2.31 $\pm$ 0.15    & 1.27 $\pm$ 0.25    & \textbf{1.08 $\pm$ 0.12}    \\
$\kappa_{max}$ & 631.89 $\pm$ 86.81 & 496.99 $\pm$ 22.02 & 583.74 $\pm$ 52.08 & 177.99 $\pm$ 13.25 & \textbf{148.16 $\pm$ 10.09} \\ \bottomrule
\end{tabular}
}
\end{table}

\clearpage

\end{document}